\documentclass[letterpaper, 10 pt, journal, twoside]{ieeetran}
\usepackage{amsmath,amsfonts}
\usepackage{algorithmic}
\usepackage{algorithm}
\usepackage{array}
\usepackage[caption=false,font=footnotesize]{subfig}
\usepackage{textcomp}
\usepackage{stfloats}
\usepackage{amssymb}
\usepackage{hyperref}
\hypersetup{
    colorlinks=true,
    linkcolor=blue,
    citecolor=blue,
    urlcolor=blue
}
\usepackage{cleveref}
\usepackage{multirow}
\usepackage{booktabs} 
\usepackage{makecell}
\usepackage{tabularx}
\usepackage{nicematrix}
\usepackage{url}
\usepackage{siunitx}
\usepackage[table]{xcolor}
\usepackage{bm}
\usepackage{verbatim}
\usepackage{graphicx}
\usepackage{cite}
\usepackage[nolist]{acronym}
\begin{document}
\pagestyle{plain}

\begin{acronym}
	\acro{OCP}{Optimal Control Problem}
	\acro{MPC}{Model Predictive Control}
	\acro{SQP}{Sequential Quadratic Programming}
	\acro{EKF}{Extended Kalman Filter}
	\acro{POV}{Point of View}
	\acro{ODE}{Ordinary Differential Equation}
	\acro{DMP}{Dynamic Motion Primitives}
    \acro{GP}{Gaussian Process}
    \acroplural{GP}[GPs]{Gaussian Processes}
    \acro{RBF}{Radial Basis Function}
    \acro{MAE}{Mean Absolut Error}
    \acro{COR}{Coefficient of Restitution}
    \acro{RK4}{Runge-Kutta 4}
    \acro{MLP}{Multi-Layer Perceptron}
    \acro{SNR}{Signal Noise Ratio}
\end{acronym}

\title{Learning Racket-Ball Bounce Dynamics Across Diverse Rubbers for Robotic Table Tennis}


\author{Thomas Gossard%
\thanks{Project page: \protect\url{https://github.com/gossardt/tt_rackets}}
}



\maketitle

\begin{abstract}
Accurate dynamic models for racket-ball bounces are essential for reliable control in robotic table tennis.
Existing models typically assume simple linear models and are restricted to inverted rubbers, limiting their ability to generalize across the wide variety of rackets encountered in practice. 
In this work, we present a unified framework for modeling ball-racket interactions across 10 racket configurations featuring different rubber types, including inverted, anti-spin, and pimpled surfaces.
Using a high-speed multi-camera setup with spin estimation, we collected a dataset of racket-ball bounces spanning a broad range of incident velocities and spins.
We show that key physical parameters governing rebound, such as the Coefficient of Restitution and tangential impulse response, vary systematically with the impact state and differ significantly across rubbers.
To capture these effects while preserving physical interpretability, we estimate the parameters of an impulse-based contact model using Gaussian Processes conditioned on the ball's incoming velocity and spin.
The resulting model provides both accurate predictions and uncertainty estimations.
Compared to the constant parameter baselines, our approach reduces post-impact velocity and spin prediction errors across all racket types, with the largest improvements observed for nonstandard rubbers.
Furthermore, the GP-based model enables online identification of racket dynamics with few observations during gameplay.
\end{abstract}


\section{Introduction}
\label{sec:introduction}

Accurate dynamic models of robotic systems are essential for reliable performance in model-based control.
Even model-free reinforcement learning methods commonly rely on simulation to accelerate policy training, where the fidelity of the simulated dynamics strongly influences the transfer of learned policies to the real system.
Consequently, inaccurate dynamics exacerbate the sim-to-real gap.

In robotic table tennis, systems can now outperform amateur players and sustain long rallies~\cite{dambrosio2024}.
However, at higher levels of play, increased ball velocities and spin magnitudes amplify the impact of modeling errors.
In particular, accurate modeling of ball-racket interactions is critical, as it directly affects both trajectory prediction and stroke planning.
Table tennis dynamics have been extensively studied in the context of developing table tennis robots~\cite{liu2012,bao2012,yang2023a,hayakawa2016,nakashima2011,zhao2017a}.
While aerodynamic flight models and table bounce dynamics are relatively well understood, the most used racket models in practice typically rely on simple linear formulations.
They are largely restricted to inverted rubbers~\cite{liu2012,nakashima2010,kyohei2019}.
However, a wide variety of blade-rubber combinations exists in practice, including pips-out rubbers, anti-spin surfaces, and blades with different stiffnesses and sponge thicknesses.

\begin{figure}[t]
    \centering
    \includegraphics[width=0.7\linewidth]{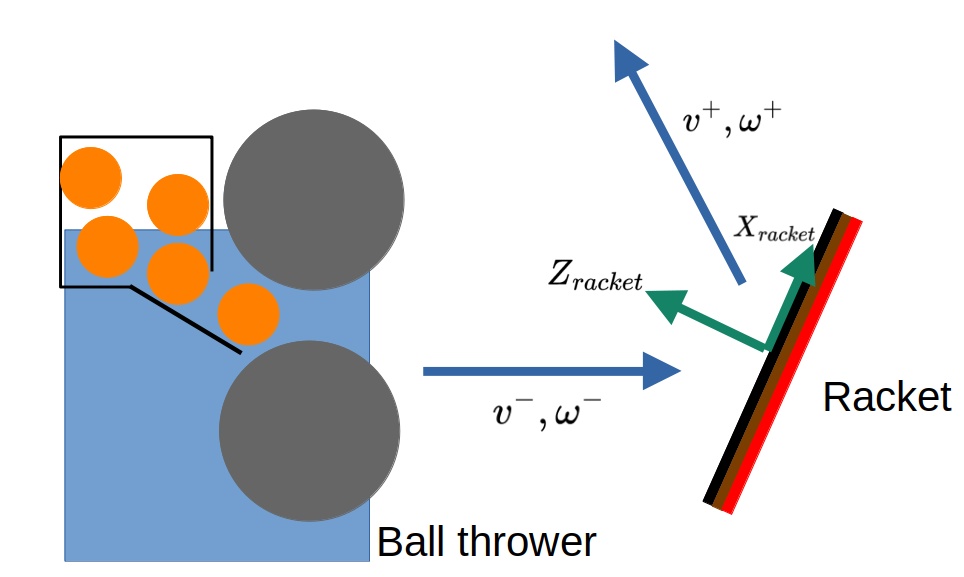}
    \caption{Experimental setup for recording racket bounces in 2D. The camera is orthogonal to the bounce plane.}
    \label{fig:ball_thrower}
\end{figure}
Studying racket-ball dynamics is challenging because the ball's high velocities and spin rates make precise measurements difficult without specialized high-speed sensing.
As a result, existing datasets are scarce, and prior work has largely relied on regressions over limited observations.
This scarcity has led to the use of simplified physical models and has limited the use of more expressive data-driven approaches.

Accurate racket-ball models are nevertheless crucial.
They enable earlier and more reliable prediction of the ball's post-impact trajectory, for example, by inferring it directly from the racket's orientation and velocity at contact rather than waiting to observe the return flight.
Such models are also essential for optimal control strategies and for reinforcement learning approaches that rely on accurate simulation.

To address these limitations, we collected a dataset of ball bounces across 10 different racket configurations, including challenging cases such as anti-spin and long-pimple rubbers.
This dataset enables the use of machine learning techniques within a Scientific Machine Learning (SciML) framework, in which data-driven models capture complex physical phenomena~\cite{quarteroni2025combining}.
However, fully data-driven approaches such as neural networks typically require large datasets, generalize poorly outside the training distribution, and offer limited interpretability.
Although our dataset is significantly larger than those used in prior works, it remains relatively small for training neural networks effectively.
Moreover, the underlying physical parameters exhibit structured dependencies on a low-dimensional input space, defined by the tangential and normal velocities $(v_s, v_z)$.
\acp{GP} are particularly well-suited to this setting, as they encode smoothness priors, perform well in data-efficient regimes, and provide uncertainty estimates.
Uncertainty quantification enables uncertainty-aware applications, such as domain adaptation in reinforcement learning~\cite{muratore2021a}.
In contrast to neural networks, \acp{GP} model smooth functions, which is consistent with the expected behavior of physical parameters.
Moreover, their Bayesian formulation enables the definition of informative priors over racket dynamics, facilitating rapid online adaptation to new opponents during a rally.
Finally, their differentiable structure makes them naturally compatible with gradient-based optimal control, which is essential for MPC-based table tennis robots~\cite{nguyen2025b}.

With this paper, we make the following contributions:
\begin{itemize}
    \item A comprehensive dataset of racket-ball interactions across 10 different racket configurations, including challenging cases such as anti-spin and long-pimple rubbers.
    
    \item A systematic and quantitative analysis of rubber-specific bounce dynamics, highlighting the strong dependence of physical parameters on the impact state and the limitations of existing constant and linear models.
    
    \item A Gaussian Process-based framework for learning state-dependent racket bounce parameters, enabling accurate modeling of nonlinear dynamics together with uncertainty quantification and online adaptation capabilities.
\end{itemize}


\section{Related Work}
\label{sec:related_work}

\textbf{Table tennis dynamic models} are the key to a high-performance table tennis robot.
Ball flight and table bounce were the first extensively studied for ball trajectory prediction.
Most existing approaches model the bounce using a linear formulation:
\begin{equation}
\begin{split}
\bm{v^{+}} & = \bm{A} \bm{v^{-}} + \bm{B} \bm{\omega^{-}}\\
\bm{\omega^{+}} & = \bm{C} \bm{v^{-}} + \bm{D} \bm{\omega^{-}}
\end{split}
\label{eq:linear_bounce_model}
\end{equation}
where $\bm{A}, \bm{B}, \bm{C}, \bm{D} \in \mathbb{R}^{3 \times 3}$.
While different models vary in the specific values of these matrices, they typically share the following structure:
\begin{equation}
	\begin{aligned}
		\bm{A}=&\left[\begin{array}{ccc}
			a_1 & 0 & 0 \\
			0 & a_1 & 0 \\
			0 & 0 & -a_2
		\end{array}\right] \quad &\bm{B}&=\left[\begin{array}{ccc}
			0 & b & 0 \\
			-b & 0 & 0 \\
			0 & 0 & 0
		\end{array}\right] \\
		\bm{C}=&\left[\begin{array}{ccc}
			0 & -c & 0 \\
			c & 0 & 0 \\
			0 & 0 & 0
		\end{array}\right] \quad &\bm{D}&=\left[\begin{array}{ccc}
			d_1 & 0 & 0 \\
			0 & d_1 & 0 \\
			0 & 0 & d_2
		\end{array}\right]
	\end{aligned}
	\label{eq:bounce_matrices}
\end{equation}
This matrix structure reflects an underlying physical assumption: the dynamics in the tangential plane (XY) are isotropic and coupled through friction, while the normal direction (Z) is decoupled.
Huang et al.~\cite{huang2011a} introduced a basic empirical linear model for ball-table bounces, offering a straightforward formulation but lacking a foundation in physical principles.
In contrast, \cite{chen2010} proposed a more physically grounded approach using a linear impulse-based model incorporating Coulomb friction to better capture tangential dynamics during impact.
It distinguishes two cases: the ball can have either rolling or sliding contact.
The nature of the contact is determined by the coefficient:
\begin{equation}
\alpha = \frac{\mu (1 + e)\,|v_z^-|}{v_s},
\quad
v_s = \left\| 
\begin{bmatrix}
v_x^- + r\,\omega_y^- \\
v_y^- + r\,\omega_x^-
\end{bmatrix}
\right\|_2
\label{eq:alpha}
\end{equation}
where $e$ is the \ac{COR}, $\mu$ is the friction coefficient between the ball and the table, $r$ is the ball's radius, and $v_s$ is the tangential velocity of the ball at the point of contact with the racket surface.
If $\alpha \geqslant 0.4$, then the velocity of the ball's contact point is 0 and the ball is rolling.
If $\alpha < 0.4$,  then the velocity of the ball's contact point is not 0 and the ball is sliding.
The resulting coefficients are shown in \Cref{tab:bounce_coefficients}, where $I= 2/3r^2m$ is the inertia of a hollow sphere and $m$ is the ball's mass.
There are only two parameters of this model, which results in a simpler parameterization than \cite{huang2011b}.
\begin{table}[]
    \centering
    \caption{Comparison of bounce model coefficients for different contact regimes.}
    \label{tab:bounce_coefficients}
    \setlength{\tabcolsep}{5pt}
    \begin{tabular}{lccc}
        \toprule
    \textbf{Coeff.} & \textbf{Table (Rolling)}~\cite{chen2010} & \textbf{Table (Sliding)}~\cite{chen2010} & \textbf{Racket}~\cite{nakashima2010} \\
        \midrule
        $a_1$ & $0.6$ & $1 - \alpha$ & $1 - \frac{k_p}{m}$ \\
        $a_2$ & $e$ & $e$ & $e$ \\
        $b$   & $0.4r$ & $\alpha r$ & $\frac{k_p r}{m}$ \\
        $c$   & $\frac{0.6}{r}$ & $\frac{3\alpha}{2r}$ & $\frac{k_p r}{I}$ \\
        $d_1$ & $0.4$ & $1 - \frac{3\alpha}{2}$ & $1 - \frac{k_p r^2}{I}$ \\
        $d_2$ & $1$ & $1$ & $1$ \\
        \bottomrule
    \end{tabular}
\end{table}
%
%
However, this table bounce model is insufficient for racket-ball interactions, as it cannot account for the observed spin inversion for certain strokes.
To address this limitation, Nakashima et al.~\cite{nakashima2010} proposed a model that captures the elastic behavior of inverted rubbers and makes the following assumption for the applied tangential impulse: $P_{xy}=-k_p\,v_s$.
The corresponding coefficients for the matrices are shown in \Cref{tab:bounce_coefficients}.
Still, such models rely on simplifying assumptions that affect their fidelity, such as the independence between the impact force and the tangential impulse.
Moreover, the proposed models were tested only for inverted rubbers, not for less common rubbers such as long-pip or antispin.
%
%
%

\textbf{Learning-based approaches}, in contrast, have demonstrated strong potential for modeling complex dynamics~\cite{pillonetto2025deep}, and have been successfully leveraged for controlling robotic systems~\cite{nagabandi2018neural,liu2024}.
However, such methods typically require large datasets, often generalize poorly outside the training distribution, and provide limited interpretability.
A more balanced approach consists of preserving the analytical model structure while learning its parameters using \acp{MLP}~\cite{zhao2016}, for instance, by estimating the coefficients of $\bm{A}, \bm{B}, \bm{C}, \bm{D}$ from data.
Nevertheless, the inductive biases of \acp{MLP} remain generic and are not aligned with the structure of physical parameters, making them less suitable for modeling physically consistent parameter variations.

\textbf{Gaussian Processes} provide a principled alternative for system identification~\cite{care2023,sarkka2019} and learning dynamical functions~\cite{turner2009system}.
They combine flexibility with explicit inductive biases through kernel design, naturally encode smoothness priors, and provide uncertainty estimates, making them particularly well-suited for modeling structured physical parameters.
%
%
The uncertainty estimates provided by \acp{GP} can be leveraged in several ways, including active learning~\cite{buisson2020actively}, domain adaptation for reinforcement learning~\cite{muratore2021a}, and the design of uncertainty-aware controllers.
\acp{GP} have been successfully integrated into control frameworks, particularly within \ac{MPC}, where they improve predictive accuracy while explicitly accounting for model uncertainty, often for safety-critical applications~\cite{hewing2019cautious,kocijan2004gaussian}.
In the context of robotic table tennis, such uncertainty estimates could be used to assess the risk associated with candidate strokes and guide safer decision-making in-game.

%

\section{Trajectory Dataset Acquisition}
\label{sec:recording_traj}
To model table tennis ball dynamics, we require its full state: position $\bm{p}$, velocity $\bm{v}$, and spin $\bm{\omega}$.
Our recording setup is shown in \Cref{fig:ball_thrower}.

Balls are launched using a ball-throwing machine with controllable initial velocity and spin, and they are directed toward a racket rigidly mounted on an industrial robot arm, enabling precise control of the incident velocity and angle while ensuring repeatable impacts.
Across all experiments, we systematically vary the ball's initial linear and angular velocities to capture a broad and representative range of table tennis dynamics.
To simplify the measurement setup, the ball motion is constrained to a plane, allowing the trajectory to be recorded with a single high-speed camera operating at 370~fps.
Ball detections are obtained using color thresholding followed by blob detection.

The ball velocity $\bm{v}$ is derived from position measurements $\bm{p}$ by fitting a linear function to the trajectory before and after the bounce, avoiding noise amplification from numerical differentiation.
To isolate the velocity change due to the bounce, gravity is compensated for as
\begin{equation}
	\tilde{\bm{p}}(t) = \bm{p}(t) - \tfrac{1}{2}\bm{g} t^2 ,
\end{equation}
where $\bm{g} = [0,\,0,\,-g]^\top$.
Over the short regression window (50~ms), drag and Magnus effects can be neglected, yielding the local linear model
\begin{equation}
	\tilde{\bm{p}}(t) = \bm{p}_0 + \bm{v}\, t ,
\end{equation}
with coefficients $\bm{p}_0$ and $\bm{v}$ estimated via least squares.
The ball's spin $\bm{\omega}$ is measured via dotted-balls using  SpinDOE~\cite{Gossard2023iros}.

The ball thrower used is limited to 12 m/s and 125 rps.
In total, the dataset contains 8{,}194 bounce events, evenly distributed across 10 racket configurations.
We split the dataset into training and test sets using an 80/20 ratio, ensuring that samples from all racket configurations are represented in both sets.

\section{Gaussian Process-Based Parameter Identification}
\label{sec:gaussian_processes}
\acp{GP} are a non-parametric Bayesian framework for modeling unknown functions~\cite{bishop2006}.
Rather than parameterizing a function explicitly, a \ac{GP} defines a distribution over functions, enabling both flexible modeling and principled uncertainty quantification.
Formally, a \ac{GP} assumes that function values at any finite set of inputs follow a joint Gaussian distribution, whose covariance structure is defined by a kernel function.

A \ac{GP} is defined as:
\begin{align}
f &\sim \mathcal{GP}\left(0, k\left(x, x^{\prime}\right)\right), \\
y &= f(x) + \epsilon, \quad \epsilon \sim \mathcal{N}(0, \sigma_n^2),
\end{align}
where $x, x' \in \mathbf{X}$ are inputs, $k$ is the covariance (kernel) function, and $y$ denotes noisy observations of the latent function $f$.
The noise term $\epsilon$ captures measurement noise and unmodeled effects, and is assumed to be independent Gaussian with variance $\sigma_n^2$.

In this work, we use the \ac{RBF} kernel:
\begin{equation}
k(x, x') = \sigma_f^2 \exp\left(-\frac{|x - x'|^2}{2l^2}\right),
\label{eq:rbf}
\end{equation}
where $\sigma_f^2$ is the signal variance and $l$ is the lengthscale, which controls the smoothness of the function.
This kernel encodes a prior assumption of smooth, continuous variations in the modeled parameters, which is consistent with the expected behavior of physical contact dynamics.

The kernel hyperparameters ${\sigma_f^2, l, \sigma_n^2}$ are learned by maximizing the marginal likelihood $p(y \mid x)$, allowing the model to automatically adapt its complexity to the data.
We optimize both hyperparameters on the training set of each racket.
This provides a flexible yet well-regularized model that captures nonlinear dependencies while maintaining uncertainty estimates.

\section{Racket Models}
\label{sec:racket_models}
In this section, we analyze and compare the dynamics of different table tennis rackets.
A table tennis racket consists of three main components: the blade, the sponge, and the rubber.
In some configurations, rubbers are used without a sponge (referred to as orthodox rubbers, e.g., Racket~6).
The blade can be made from wood (e.g., balsa, limba, cypress, hinoki) or composite materials such as carbon fiber, with its material, thickness, and number of plies influencing stiffness, weight, and overall playing characteristics.
The sponge thickness strongly affects energy transfer during impact, with thicker sponges generally enabling higher energy restitution.

Rubbers are classified into four primary categories, each offering distinct playing characteristics: anti-spin rubbers, inverted rubbers (pimples-in), short-pimple rubbers, and long-pimple rubbers.
We collected a variety of rackets to examine the impact of different blade and rubber combinations on the racket's bounce dynamics.
The specific blades and rubbers used are listed in \Cref{tab:rubber} and \Cref{tab:blades}.
We selected three blade types, Offensive, Allround, and Defensive, and combined them with representative rubber types to ensure broad coverage.
This resulted in ten racket configurations, detailed in \Cref{tab:racket_combinations}, which are analyzed by rubber type.
\begin{table}[]
\centering
\rowcolors{1}{gray!25}{}
\begin{tabular}{|l|l|}
\hline
\textbf{Name}              & \textbf{Type}        \\
\hline
Tibhar Grass D.TecS        & Long pips            \\
Dr. Neubauer Diamant       & Medium pips          \\
andro Blowfish             & Short pips           \\
andro Power 3              & Inverted (allround)  \\
andro Hexer Duro           & Inverted (offensive) \\
Dr. Neubauer A-B-S II soft & Anti-spin            \\
\hline
\end{tabular}
\vspace{2mm}
\caption{Rubbers used in the experiments}
\label{tab:rubber}
\end{table}

\begin{table}[]
\centering
\rowcolors{1}{gray!25}{}
\begin{tabular}{|l|l|}
\hline
\textbf{Name} & \textbf{Type} \\
\hline
DONIC Appelgren Allplay Senso V1 & Allrounder \\
Donic Holz Original Carbospeed   & Offensive  \\
Tibhar Holz Defense Plus         & Defensive  \\
\hline
\end{tabular}
\vspace{2mm}
\caption{Blades used in the experiments}
\label{tab:blades}
\end{table}
\begin{table}[]
	\centering
	\begin{tabular}{|c|c|c|c|}
		\hline
		\rowcolor{gray!25}\textbf{Blade} & \textbf{Sponge thickness (mm)} & \textbf{Rubber} & \textbf{Id} \\
		\hline
		\multirow{2}{*}{Offensive} & 2.1 & Inverted (offensive) & 1 \\
		& \cellcolor{gray!25}1.8 & \cellcolor{gray!25}Inverted (allround) & \cellcolor{gray!25}2 \\
		\hline
		\multirow{2}{*}{Defensive} & 2.1 & Inverted (offensive) & 3 \\
		& \cellcolor{gray!25}1.8 & \cellcolor{gray!25}Inverted (allround) & \cellcolor{gray!25}4 \\
		\hline
		\multirow{6}{*}{Allrounder} & 1.2 & Long pips & 5 \\
		& \cellcolor{gray!25}0   & \cellcolor{gray!25}Long pips & \cellcolor{gray!25}6 \\
		& 1.2 & Medium pips & 7 \\
		& \cellcolor{gray!25}2.0 & \cellcolor{gray!25}Short pips & \cellcolor{gray!25}8 \\
		& 2.1 & Inverted (offensive) & 9 \\
		& \cellcolor{gray!25}2.1 & \cellcolor{gray!25}Anti-spin & \cellcolor{gray!25}10 \\
		\hline
	\end{tabular}
    \vspace{2mm}
	\caption{Racket configurations: Components used to build the different rackets.}
	\label{tab:racket_combinations}
\end{table}




\subsection{Anti-spin rubber}
\label{ssec:antispin}
\begin{figure}[]
    \centering
    \subfloat[Coefficient of restitution (COR)]{%
        \includegraphics[width=0.9\linewidth]{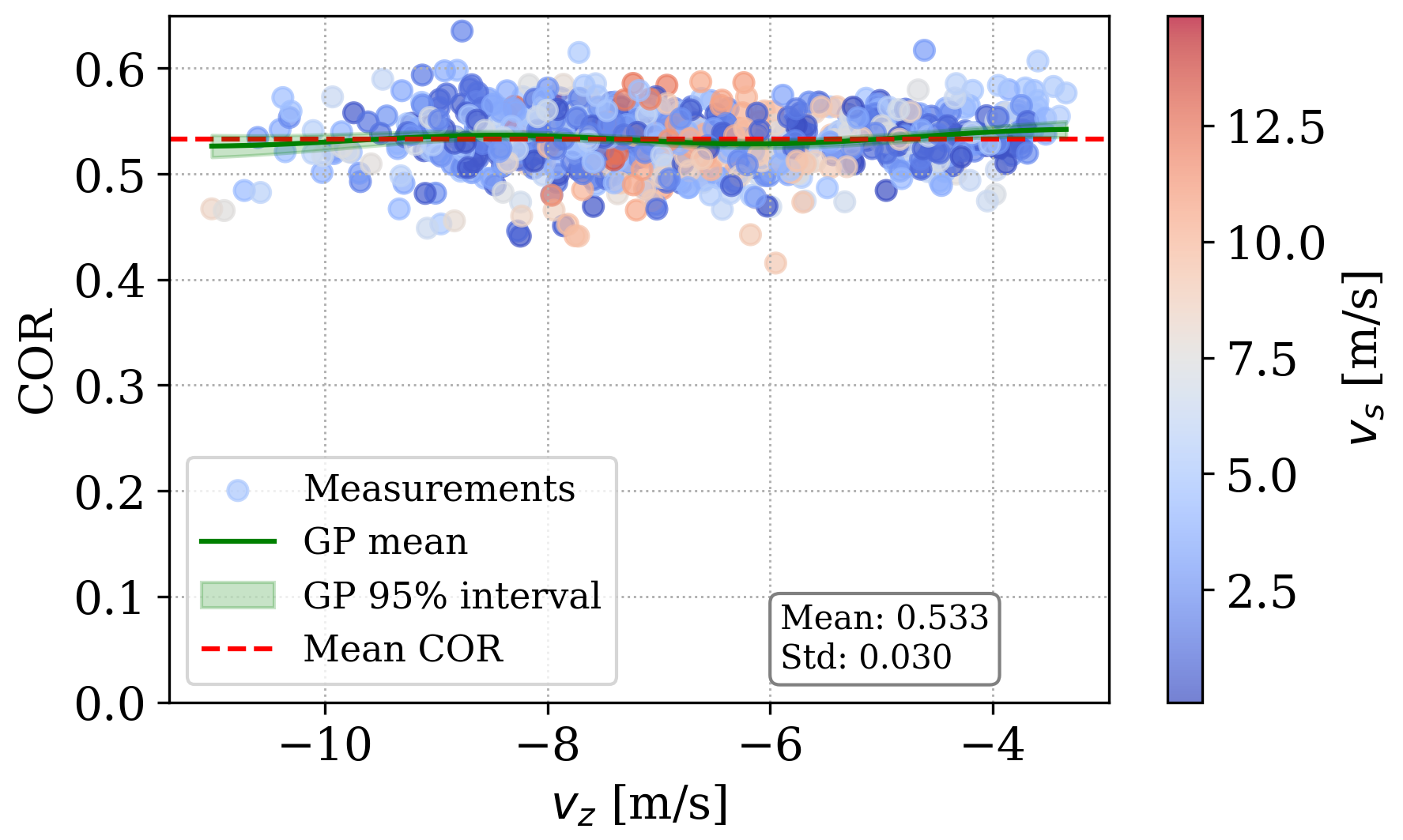}
    }
    \hfill
    \subfloat[Coefficient $\alpha$]{%
        \includegraphics[width=0.9\linewidth]{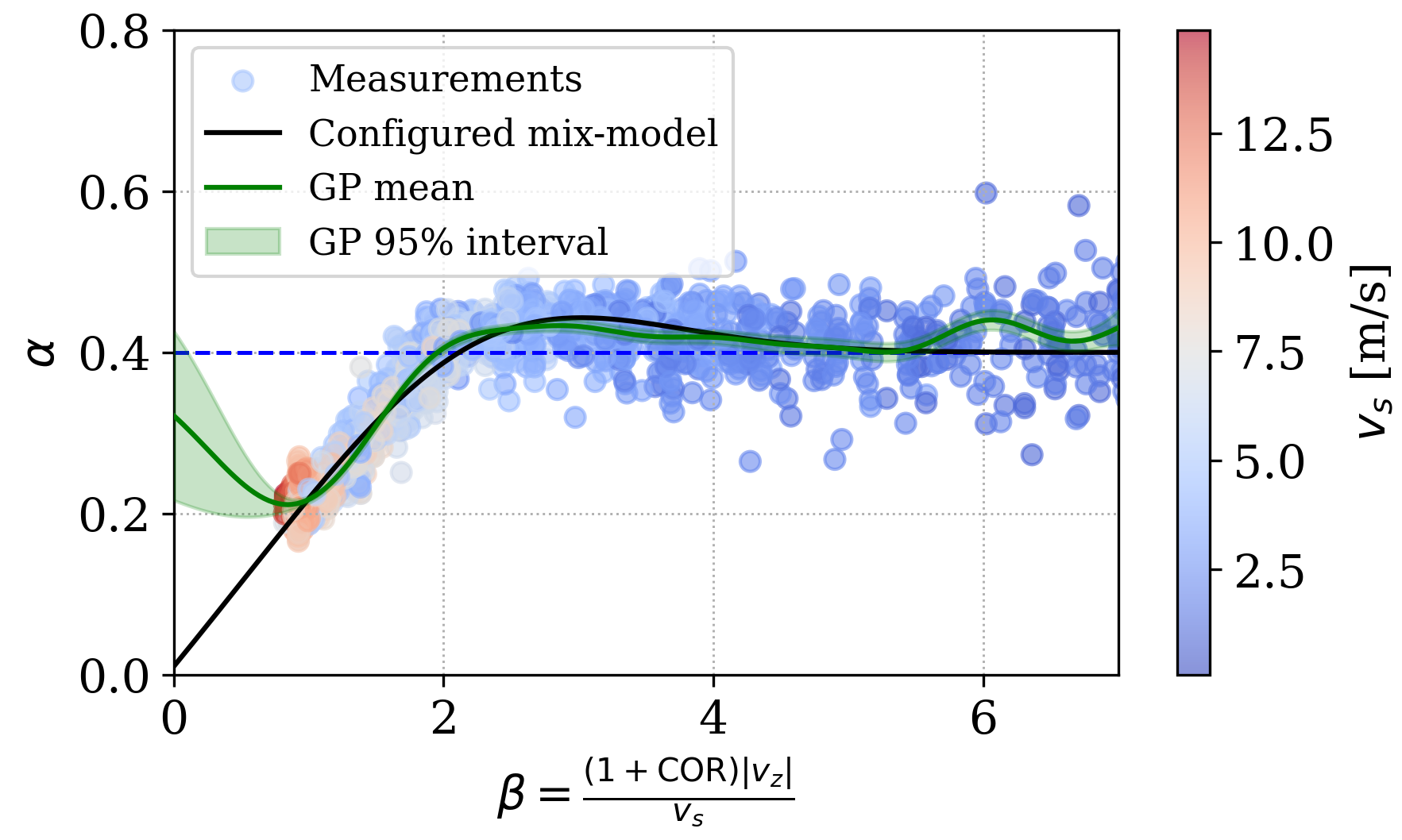}
    }
    \caption{Bounce dynamics of the anti-spin racket (Racket 10)}
    \label{fig:racket_10}
\end{figure}

Anti-spin rubbers are characterized by their low-friction surface, specifically designed to neutralize incoming spin rather than generate it.
During contact, the ball tends to slide across the surface, which dampens or even reverses the opponent's spin.
Given their sliding characteristics, we can model the bounce using the Coulomb friction model, analogous to the standard table bounce model.
In \Cref{fig:racket_10}, we plot $\alpha$ with respect to $(1+COR)|v_z|/v_s$ for the antispin rubber racket (Racket 10).
From \Cref{eq:linear_bounce_model} and \Cref{eq:bounce_matrices} for the sliding bounce, we can compute $\alpha$ in two distinct ways: one based on the change in linear velocity, and the other based on the change in angular velocity:
\begin{equation}
\alpha^{(v)} = \frac{v_x' - v_x}{r\,\omega_y - v_x}
\quad \text{or} \quad
\alpha^{(\omega)} = \frac{2(\omega_y' - \omega_y)}{3\left(\frac{v_x}{r} - \omega_y\right)}
\end{equation}
While both expressions are theoretically equivalent, they rely on different measured quantities, which accounts for the discrepancy in empirical results.
The estimated $\alpha$ is valid for both the sliding and rolling regimes.
In theory, we expect to observe a linear increase in $\alpha$ with a slope equal to the friction coefficient $\mu$ during sliding, followed by a plateau at 0.4 once the rolling condition is achieved.
However, fitting a \ac{GP} to the data reveals that $\alpha$ initially exceeds 0.4 before converging to this value.
This behavior can be attributed to the simplicity of the Coulomb friction model, which assumes an abrupt transition between sliding and rolling.
Additionally, the model assumes a constant friction coefficient and does not differentiate between static and kinetic friction, which have distinct values.
To address this limitation, we introduce a parametric model with a smooth transition between sliding and rolling, governed by the normal cumulative distribution function.
The parameter $\alpha$ is given by
\begin{equation}
\begin{aligned}
\alpha(\beta) =& \Phi_{\theta, \sigma}(\beta)\,\underbrace{\mu \beta}_{\text{sliding}} 
+ \left(1 - \Phi_{\theta, \sigma}(\beta)\right)\,\underbrace{0.4}_{\text{rolling}} \\
&\text{with} \quad \beta = \frac{(1 + e)\,|v_z|}{v_s}
\end{aligned}
\end{equation}
where $\Phi_{\theta, \sigma}$ denotes the cumulative density function of a normal distribution with mean $\theta$ and a standard deviation $\sigma$.
We optimized $\theta, \sigma, \mu$ to minimize the velocity prediction \ac{MAE}.
The optimal values were $\theta=2.42$, $\sigma=1.348$, yielding a friction coefficient $\mu=0.197$.
In comparison, optimizing the standard model for $\mu$ results in a friction coefficient of 0.207.

Anti-spin rubbers are additionally characterized by a low rebound speed, reflected in a low \ac{COR}.
In \Cref{fig:racket_10}, Racket~10 displays the lowest \ac{COR} ($e=0.53$), which enhances control and favors defensive strokes such as blocks and controlled pushes.

\subsection{Inverted Rubber}
\label{ssec:inv_rubber}
Inverted rubbers feature a smooth, high-friction surface that enables strong spin generation and precise control, making them the most widely used and versatile option.

\begin{figure*}[t]
\centering

\subfloat[Off. blade, Off. rubber (1)]{%
    \includegraphics[width=0.45\linewidth]{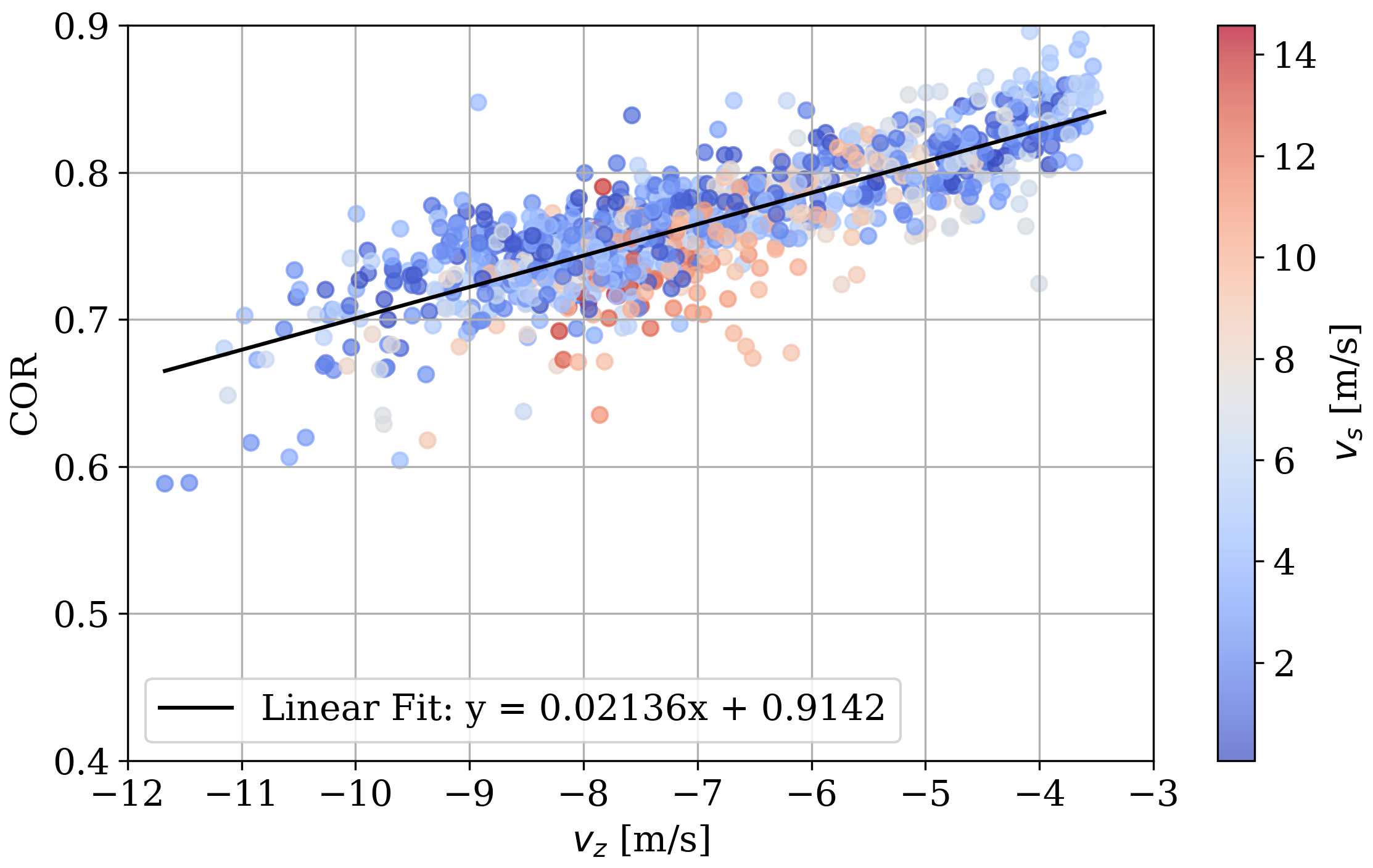}
}
\hfill
\subfloat[Off. blade, Allr. rubber (2)]{%
    \includegraphics[width=0.45\linewidth]{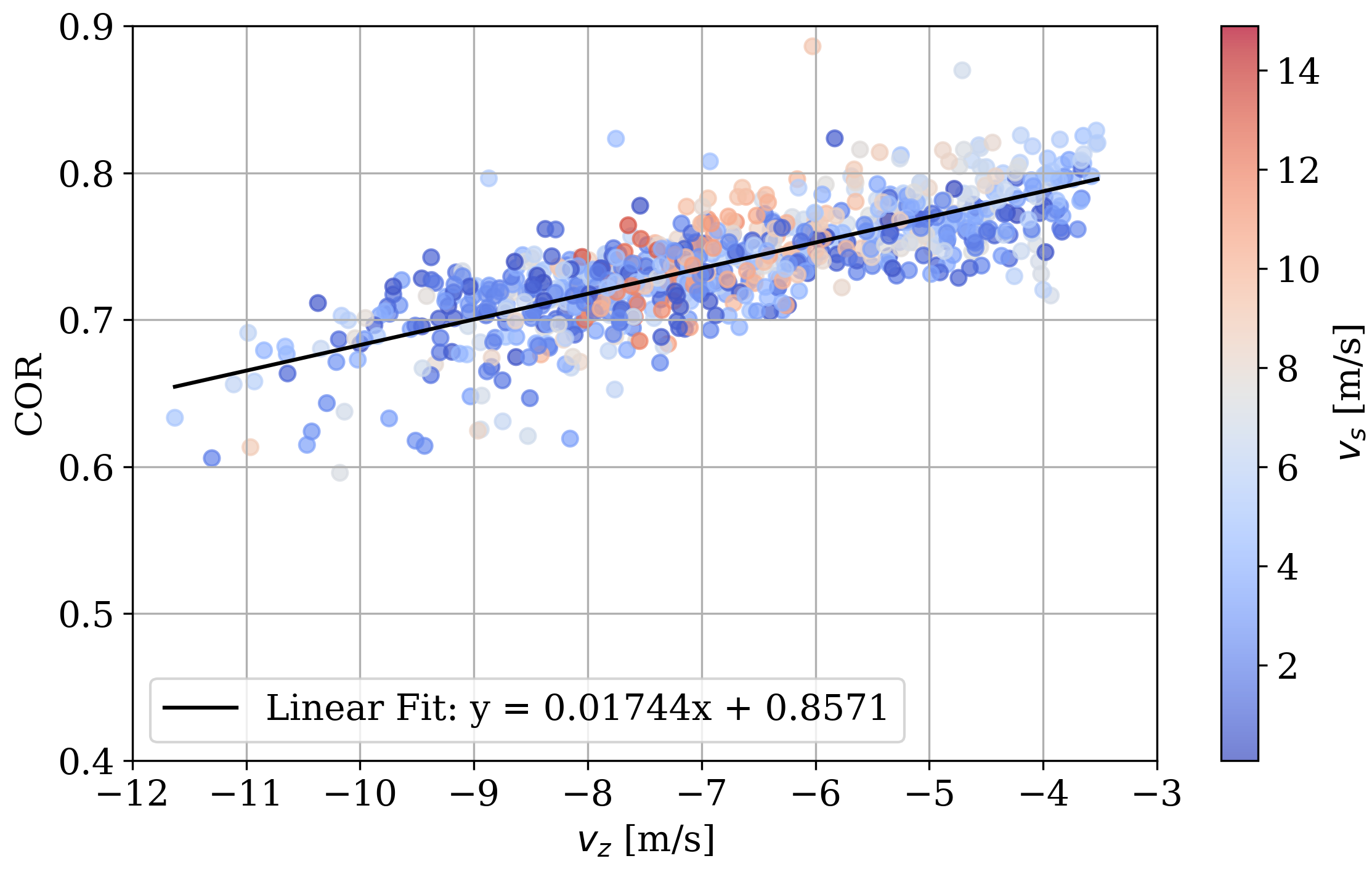}
}

\subfloat[Def. blade, Off. rubber (3)]{%
    \includegraphics[width=0.45\linewidth]{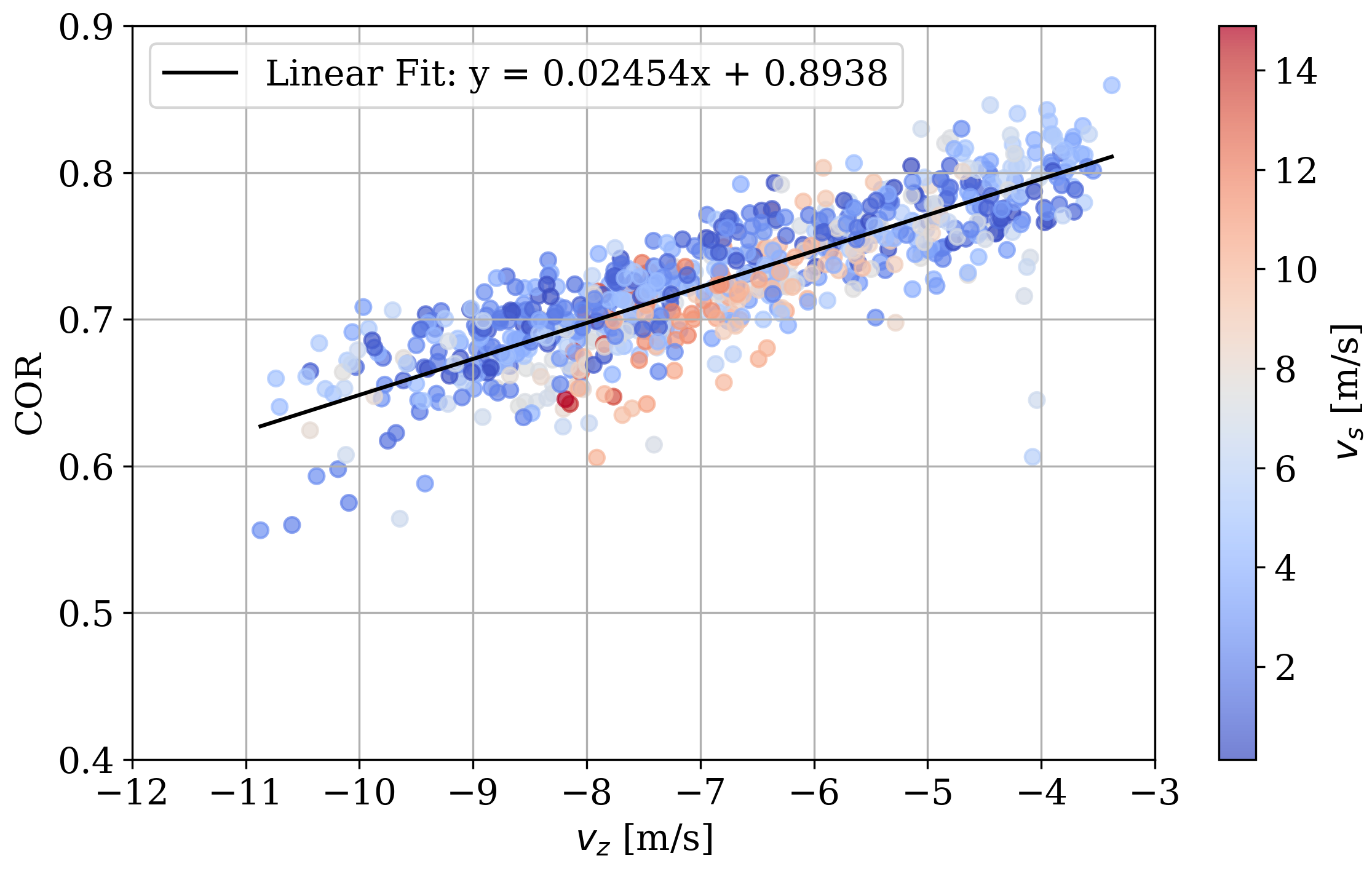}
}
\hfill
\subfloat[Def. blade, Allr. rubber (4)]{%
    \includegraphics[width=0.45\linewidth]{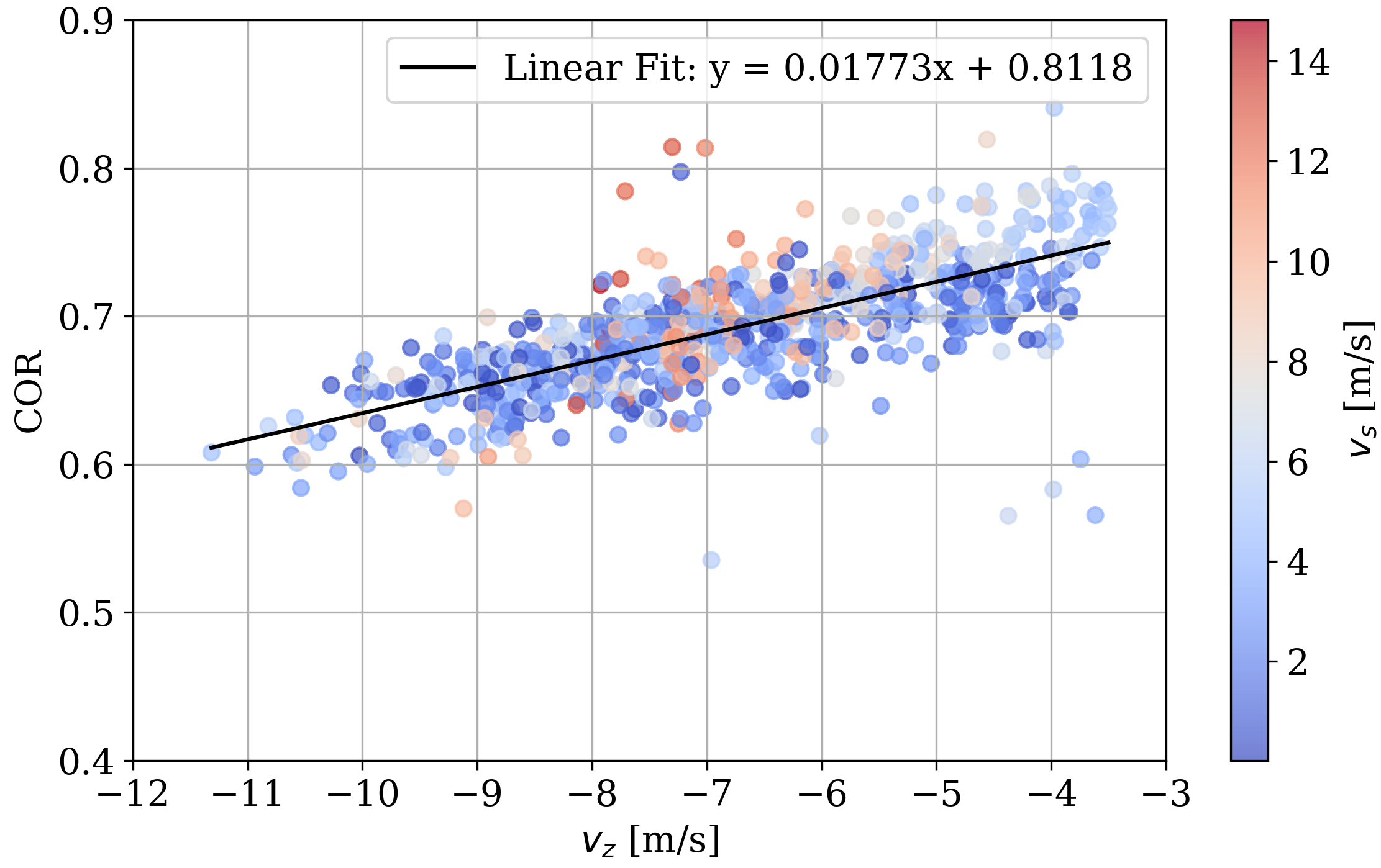}
}

\subfloat[Allr. blade, Off. rubber (9)]{%
    \includegraphics[width=0.45\linewidth]{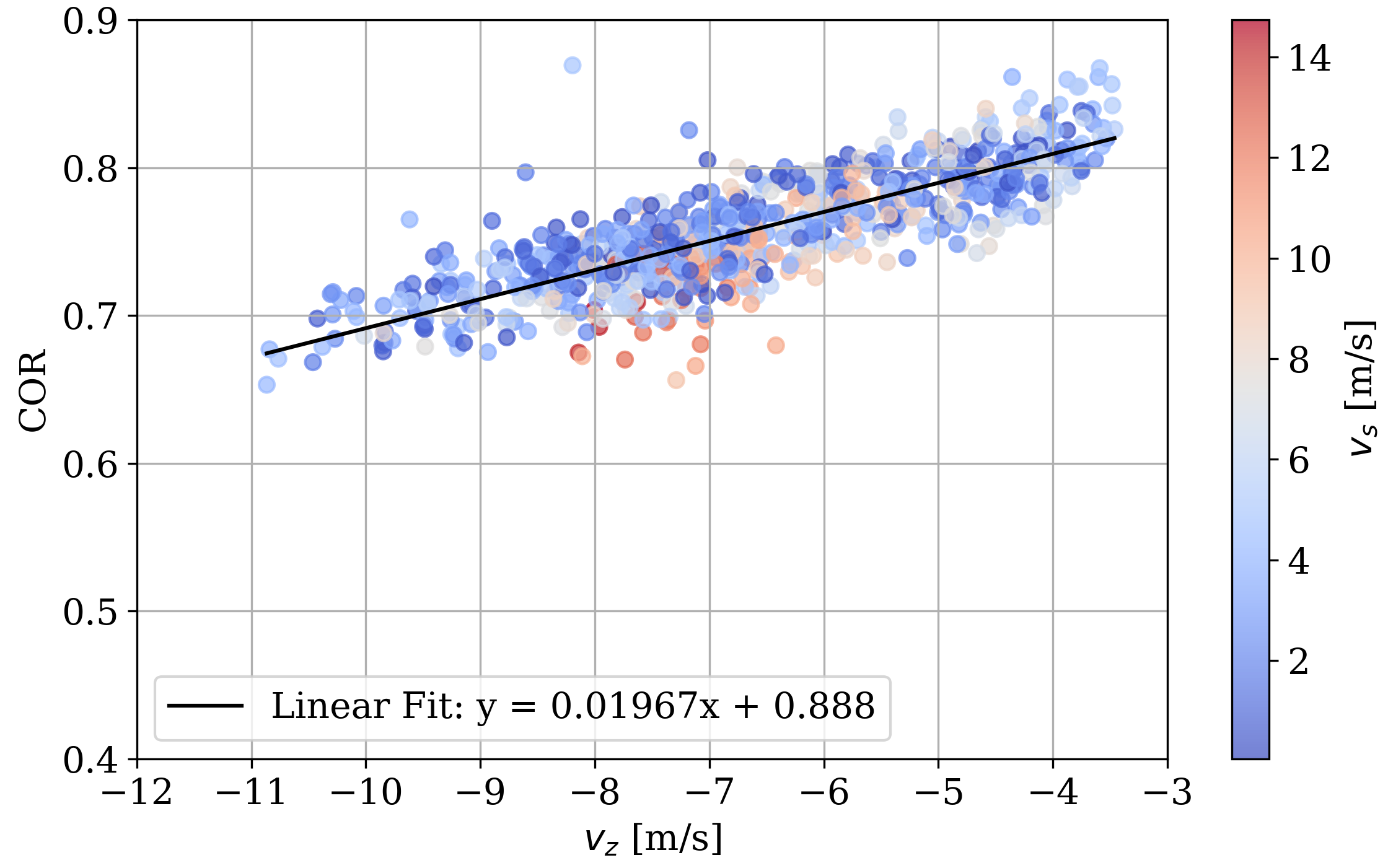}
}
\hfill
\subfloat[Fitted linear models]{%
    \includegraphics[width=0.45\linewidth]{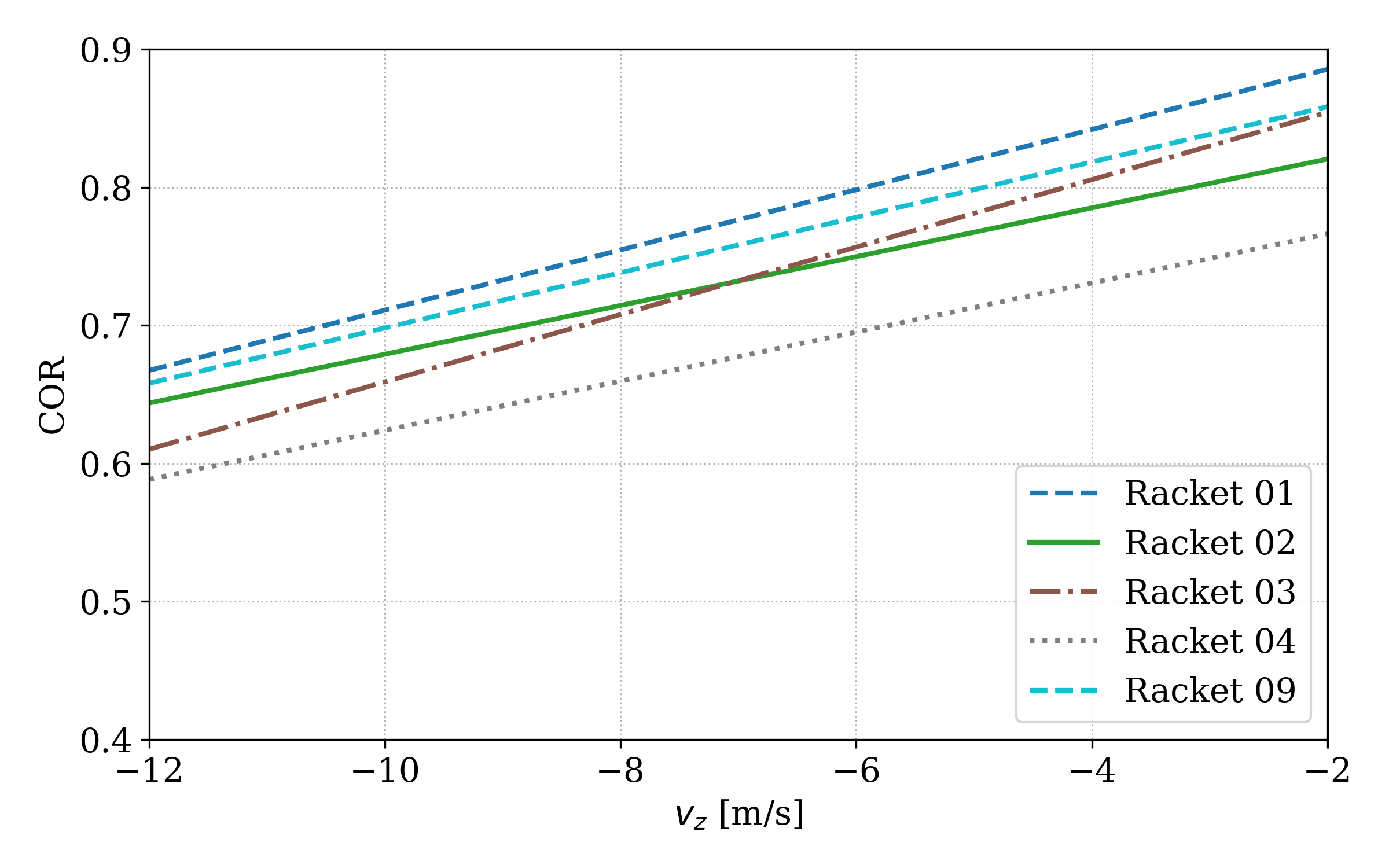}
}

\caption{COR of rackets equipped with inverted rubbers as a function of the incident normal velocity. 
The number in parentheses denotes the racket ID. The bottom-right plot shows fitted linear models for all rackets for comparison.}
\label{fig:cor_inv_rubbers}
\end{figure*}

\Cref{fig:cor_inv_rubbers} shows the variation of the \ac{COR} with respect to the normal impact velocity $v_z$ for the 5 different inverted rubber rackets (Rackets 1, 2, 3, 4, and 9).
The \ac{COR} is not constant but decreases approximately linearly with increasing $|v_z|$, indicating greater energy dissipation at higher impact speeds.
This effect is significant, with differences of up to 0.15 between low ($\sim 2$~m/s) and high ($\sim 12$~m/s) velocities.

The results also highlight the distinct roles of blade and rubber.
Rackets~1, 3, and 9 share the same offensive rubber but differ in blade type (offensive, defensive, allround), with stiffer blades yielding higher \ac{COR} due to improved energy restitution.
Rubber properties further influence the rebound: offensive rubbers (Rackets~1, 3, 9) exhibit higher \ac{COR} than all-round rubbers (Rackets~2, 4), likely due to thicker sponges that store and release more energy.

Finally, different rubbers exhibit distinct linear trends.
Offensive rubbers show a consistent slope of $0.021$ on average, while allround rubbers exhibit a shallower slope of $0.017$.

\begin{figure*}[]
\centering

\subfloat[Off. blade, Off. rubber (1)]{%
    \includegraphics[width=0.45\linewidth]{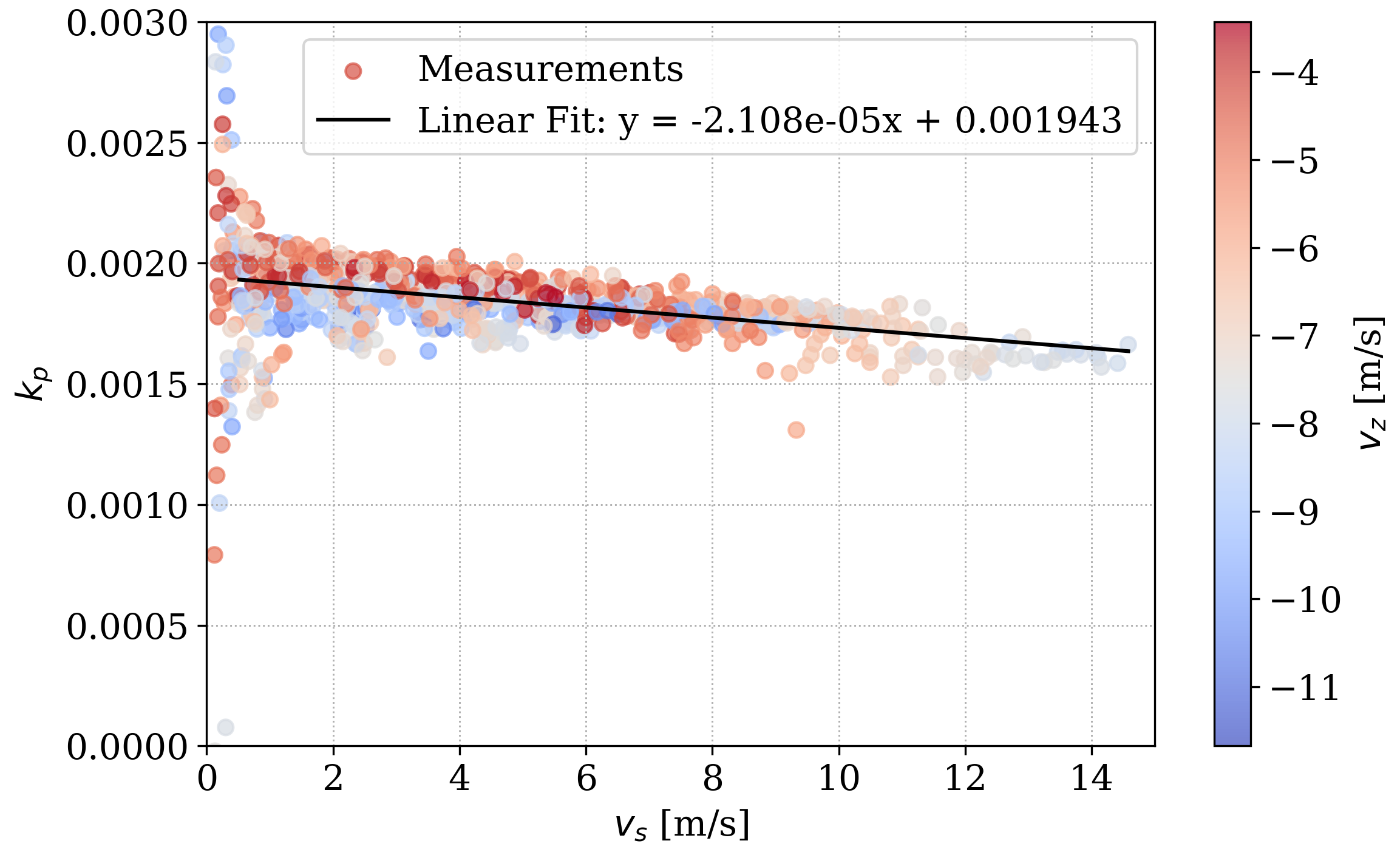}
}
\hfill
\subfloat[Off. blade, Allr. rubber (2)]{%
    \includegraphics[width=0.45\linewidth]{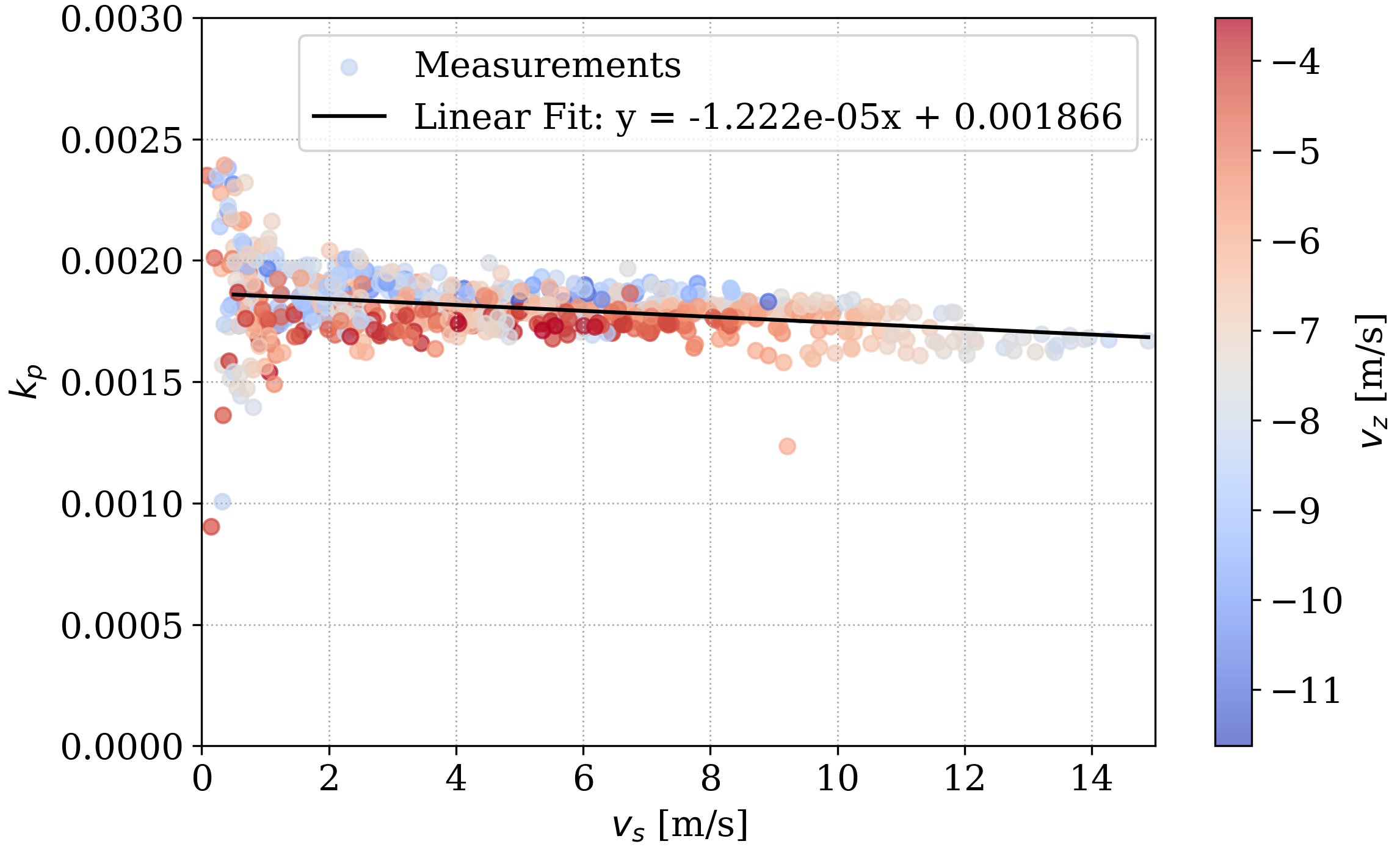}
}

\subfloat[Def. blade, Off. rubber (3)]{%
    \includegraphics[width=0.45\linewidth]{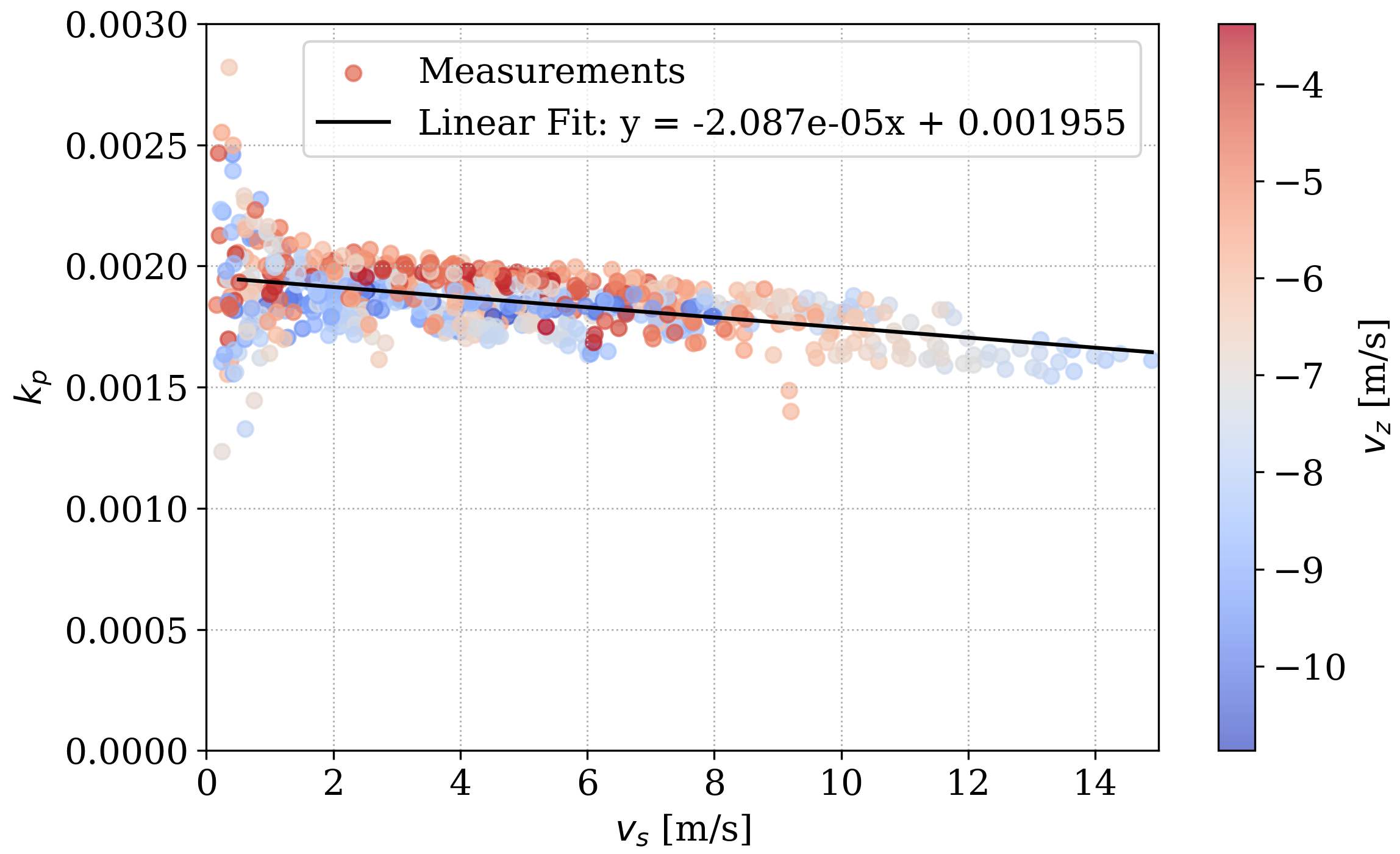}
}
\hfill
\subfloat[Def. blade, Allr. rubber (4)]{%
    \includegraphics[width=0.45\linewidth]{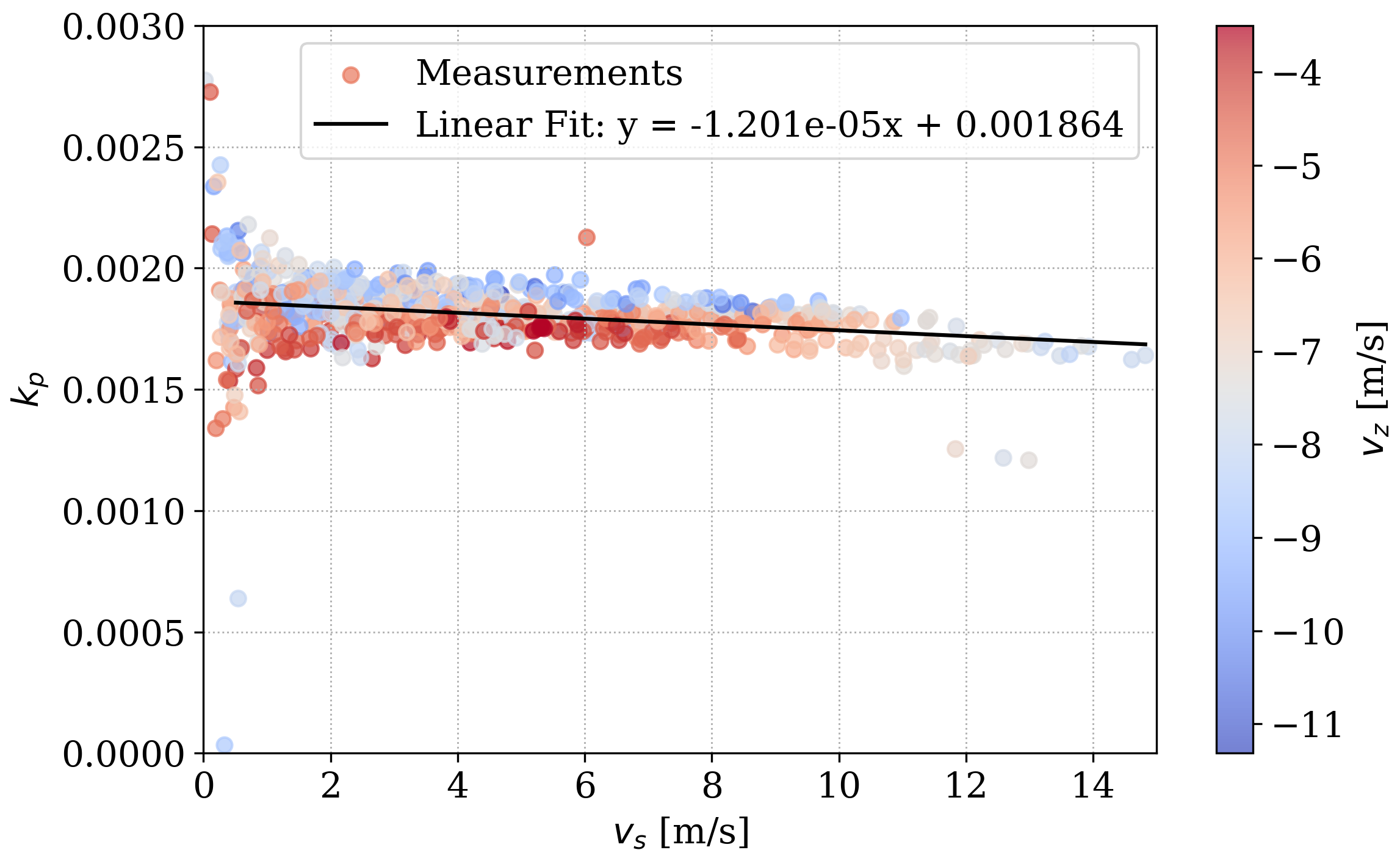}
}

\subfloat[Allr. blade, Off. rubber (9)]{%
    \includegraphics[width=0.45\linewidth]{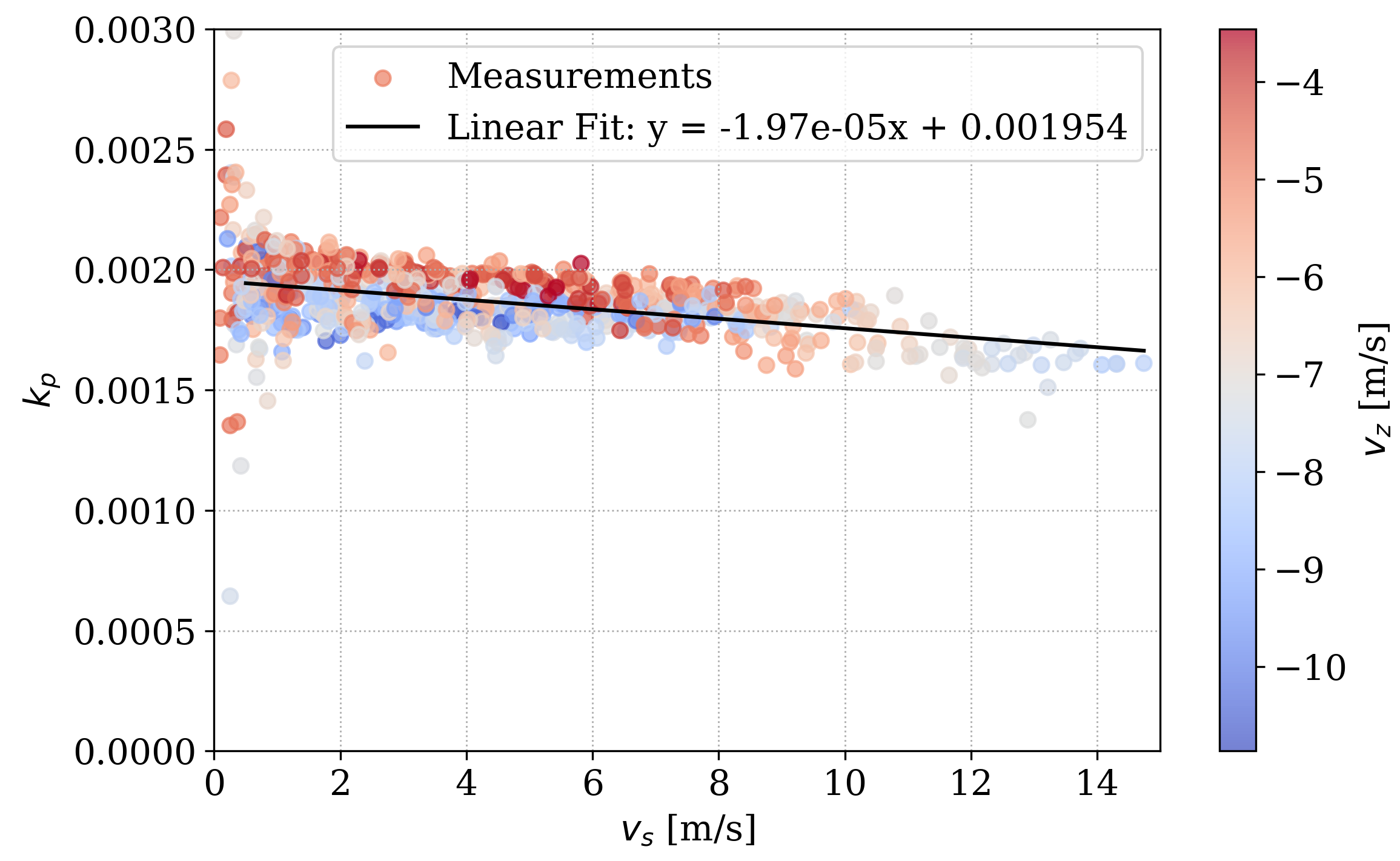}
}
\hfill
\subfloat[Fitted linear models]{%
    \includegraphics[width=0.45\linewidth]{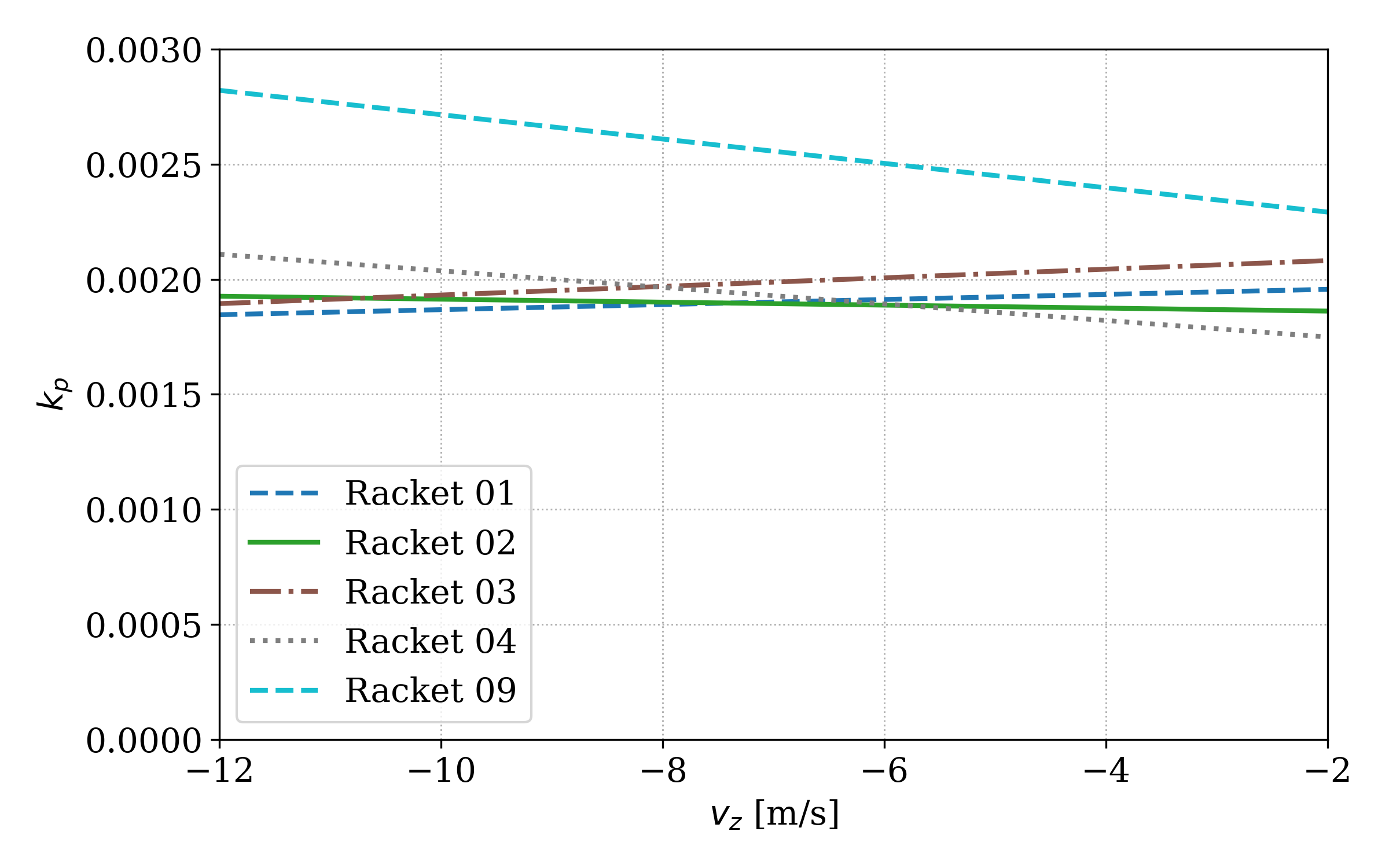}
}

\caption{$k_p$ of rackets equipped with inverted rubbers as a function of the incident normal velocity.
The bottom row shows the fitted linear model for all rackets for comparison.}
\label{fig:kp_inv_rubbers}
\end{figure*}

Analogous to $\alpha$, the parameter $k_p$ in the model of \cite{nakashima2010} can be estimated using two equivalent expressions derived from \Cref{eq:linear_bounce_model} and \Cref{tab:bounce_coefficients}:
\begin{equation}
\begin{aligned}
k_p^{(v)} &= m \,\frac{v_x' - v_x}{r\,\omega_y - v_x}
\quad \text{or} \quad
k_p^{(\omega)} = I \,\frac{\omega_y' - \omega_y}{r v_x - r^2 \omega_y}
\end{aligned}
\end{equation}
The results are shown in \Cref{fig:kp_inv_rubbers}.
As expected, the blade has minimal influence on $k_p$, with all rackets showing similar average values around $k_p \sim 0.019$, although minor variations remain.

We observe that $k_p$ depends on both the surface velocity $v_s$ and the normal velocity $v_z$.
For all inverted rubbers, $k_p$ consistently decreases with increasing $v_s$.
In contrast, its dependence on $v_z$ varies with rubber type: it increases with $|v_z|$ for all-around rubbers (Rackets~2 and 4) and decreases for offensive rubbers (Rackets~1, 3, and 9).
At low $v_s$, the signal-to-noise ratio decreases, resulting in noisier estimates.
Interestingly, the offensive rubber exhibits a higher $k_p$ at low $|v_z|$, whereas the allrounder rubber shows higher $k_p$ values at high $v_z$.



\subsection{Pips}
\label{ssec:pips}
Pimpled rubbers are less common than inverted rubbers but are valued for their disruptive behavior and reduced sensitivity to incoming spin.
Short pips produce relatively stable returns suited for counter-play, while long pips introduce more flexible, nonlinear interactions that generate spin reversal and unpredictable trajectories, particularly in sponge-less configurations (Racket 6).

\begin{figure*}[]
\centering

\subfloat[Long pips (5)]{%
    \includegraphics[width=0.48\linewidth]{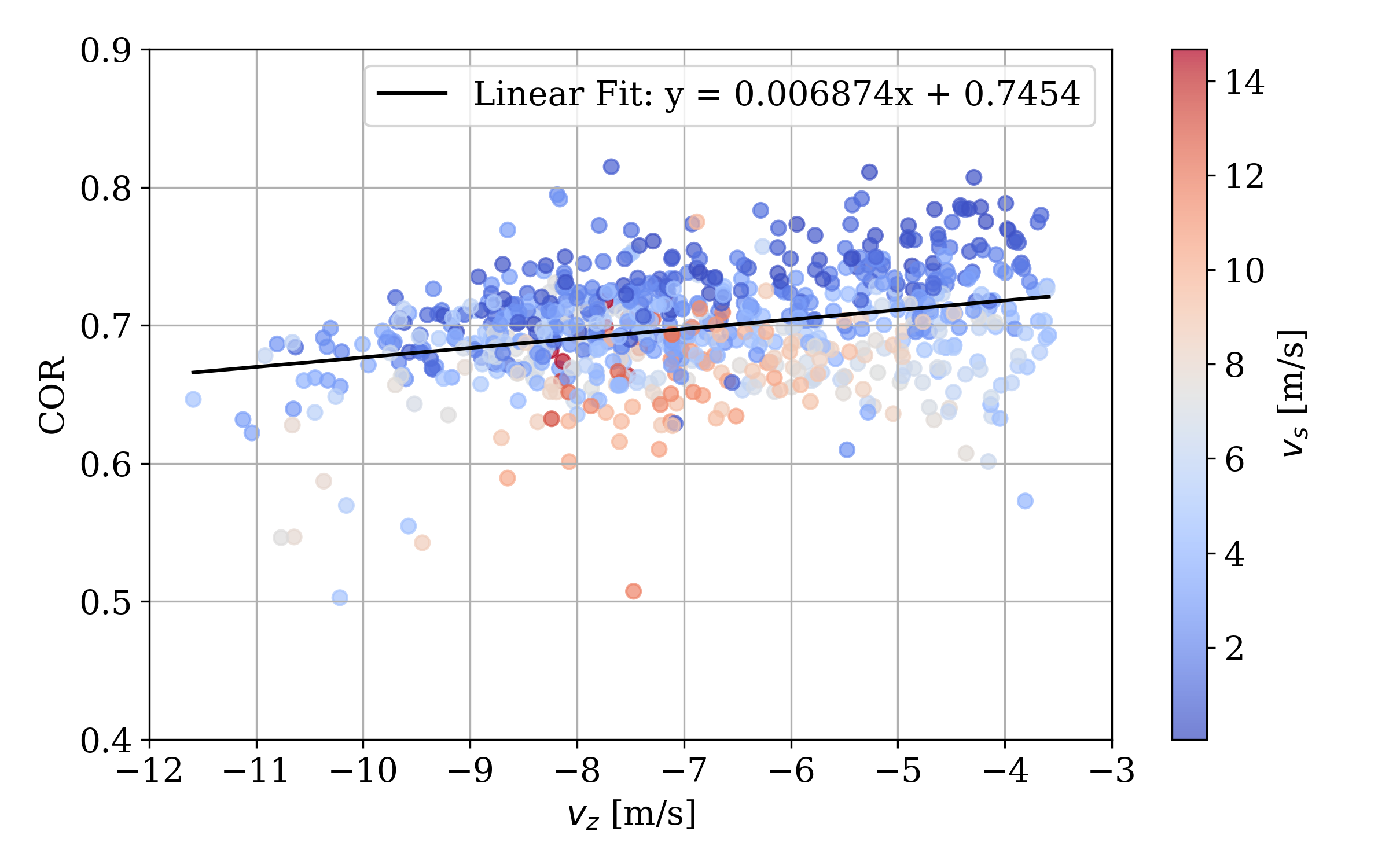}
}
\hfill
\subfloat[Long pips,no sponge (6)]{%
    \includegraphics[width=0.48\linewidth]{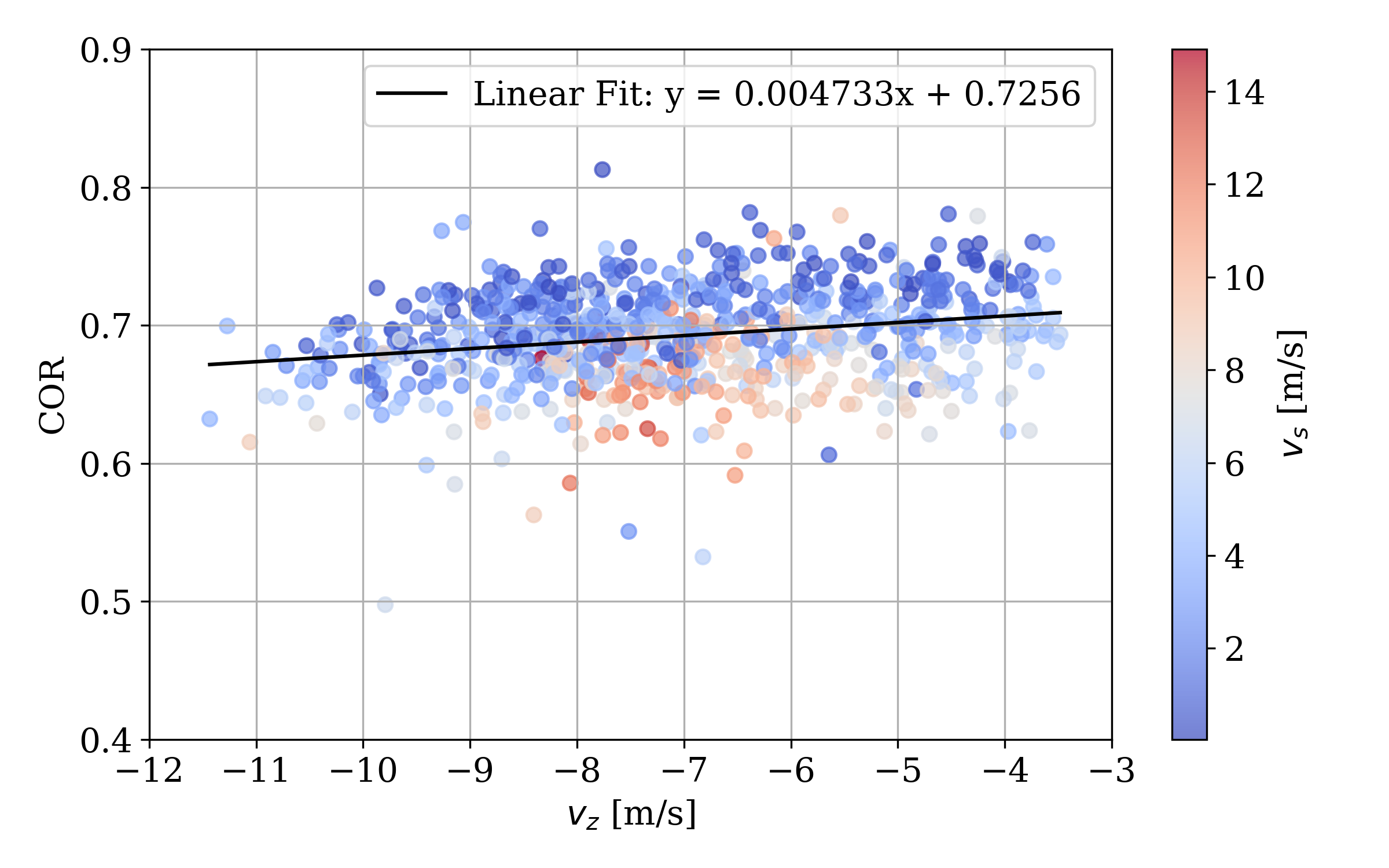}
}
\hfill
\subfloat[Medium pips (7)]{%
    \includegraphics[width=0.48\linewidth]{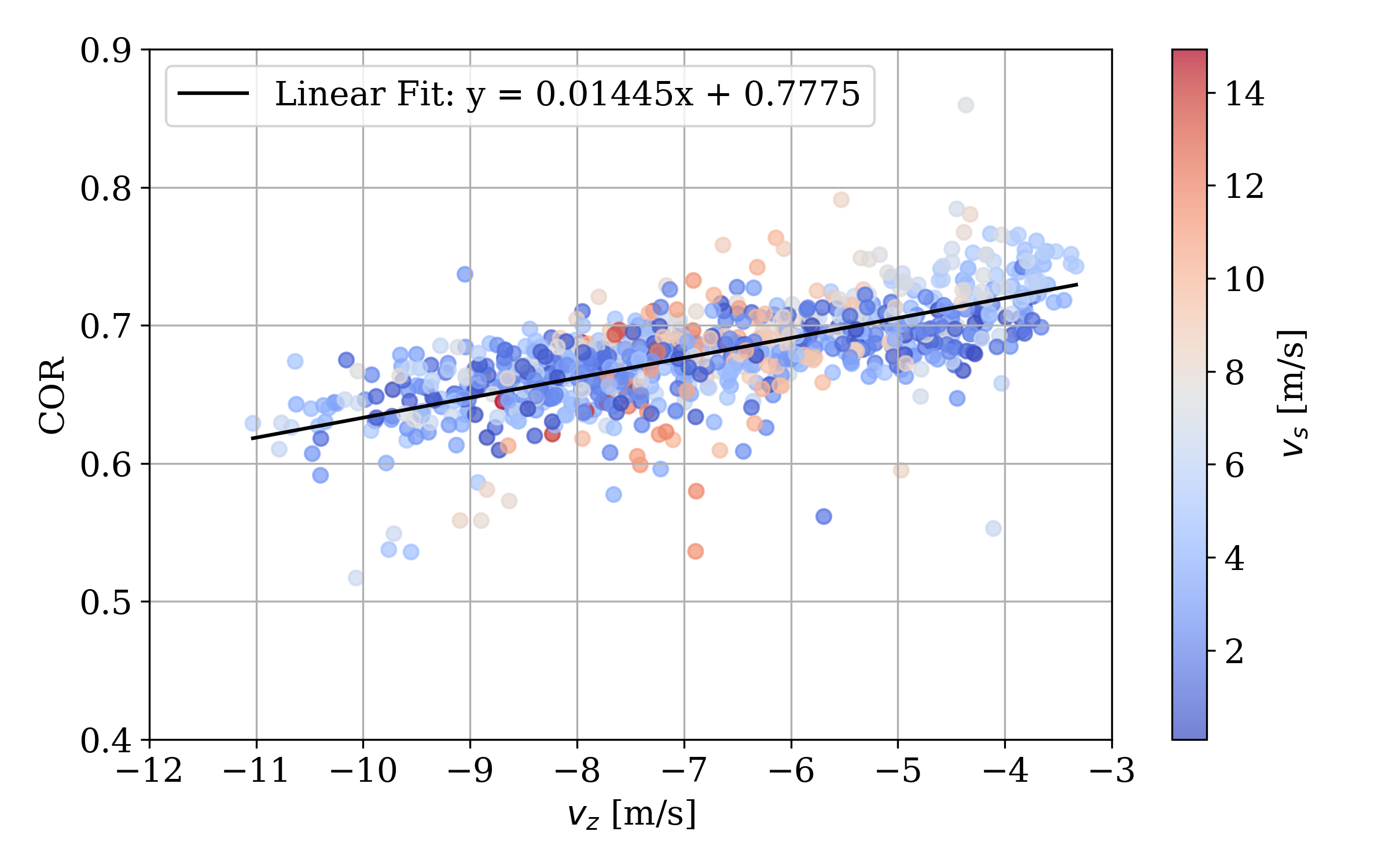}
}
\hfill
\subfloat[Short pips (8)]{%
    \includegraphics[width=0.48\linewidth]{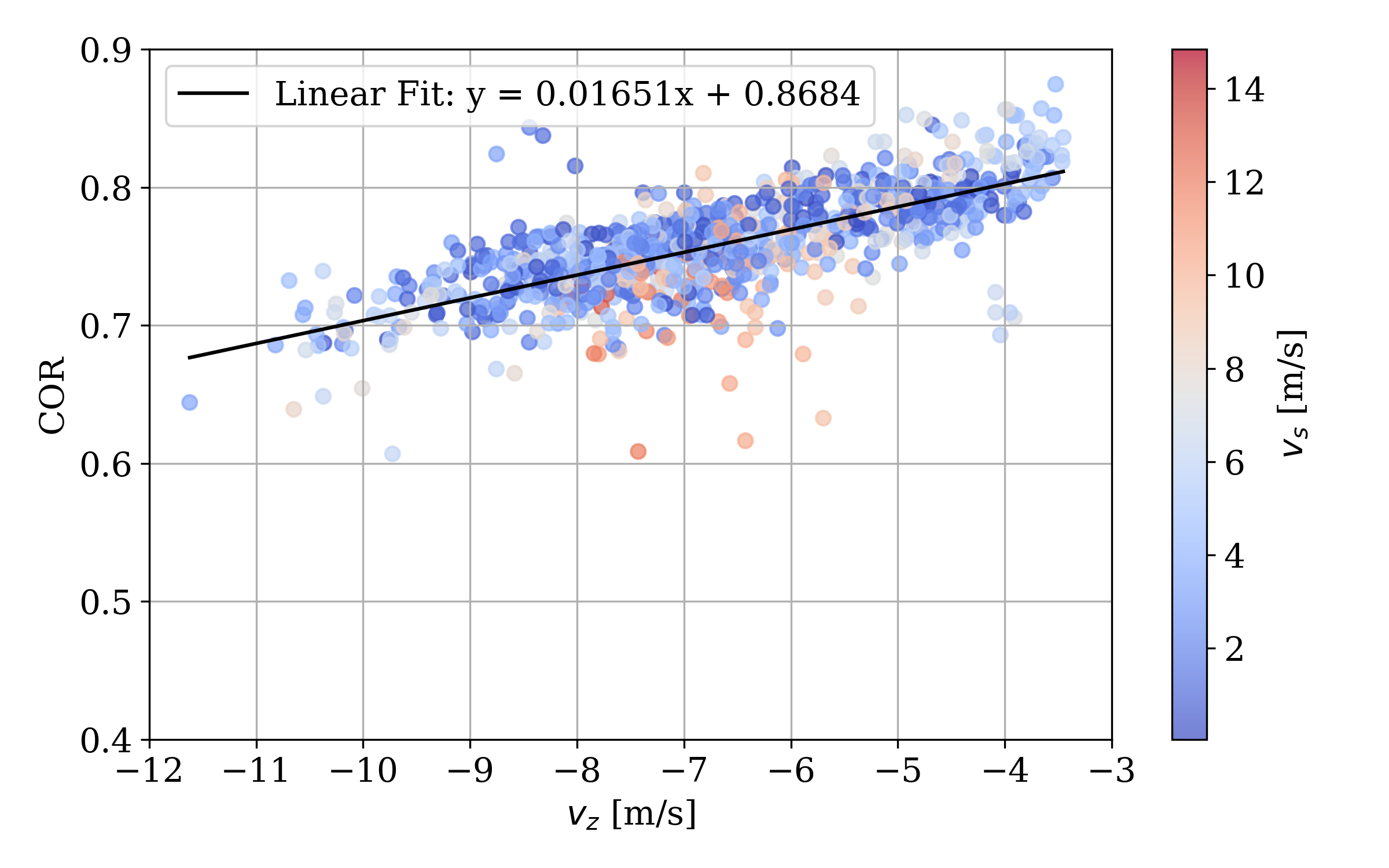}
}

\caption{Coefficient of restitution (COR) of rackets equipped with pimpled rubbers as a function of the incident normal velocity.}
\label{fig:cor_pip_rubbers}
\end{figure*}

In \Cref{fig:cor_pip_rubbers}, we plot the \ac{COR} for the pip rubbers.
For long pips (rackets 5 and 6), the \ac{COR} primarily depends on the surface velocity $v_s$, with higher $v_s$ leading to lower \ac{COR}.
This is due to the pips bending fully in one direction, reducing their ability to restore energy to the ball.
Normal velocity $v_z$ also plays a role: at higher $|v_z|$, the pips are more compressed, leading to reduced variability in \ac{COR} and more stable responses.
This behavior is specific to long pips.
Medium and short pips (rackets 7 and 8) do not exhibit the same sensitivity.
%

\begin{figure*}[]
\centering

\subfloat[Long pips (5)]{%
    \includegraphics[width=0.48\linewidth]{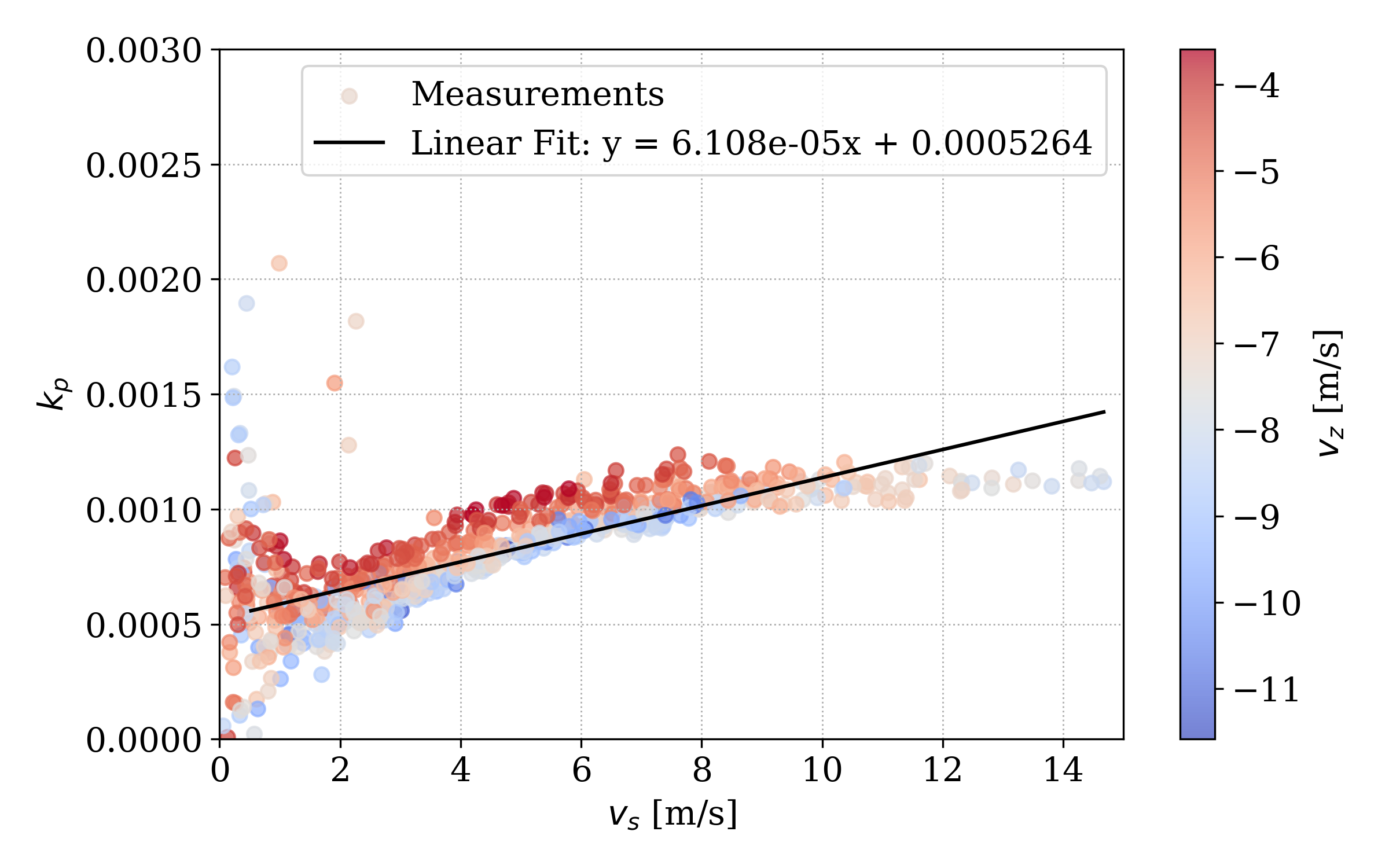}
}
\hfill
\subfloat[Long pips, no sponge (6)]{%
    \includegraphics[width=0.48\linewidth]{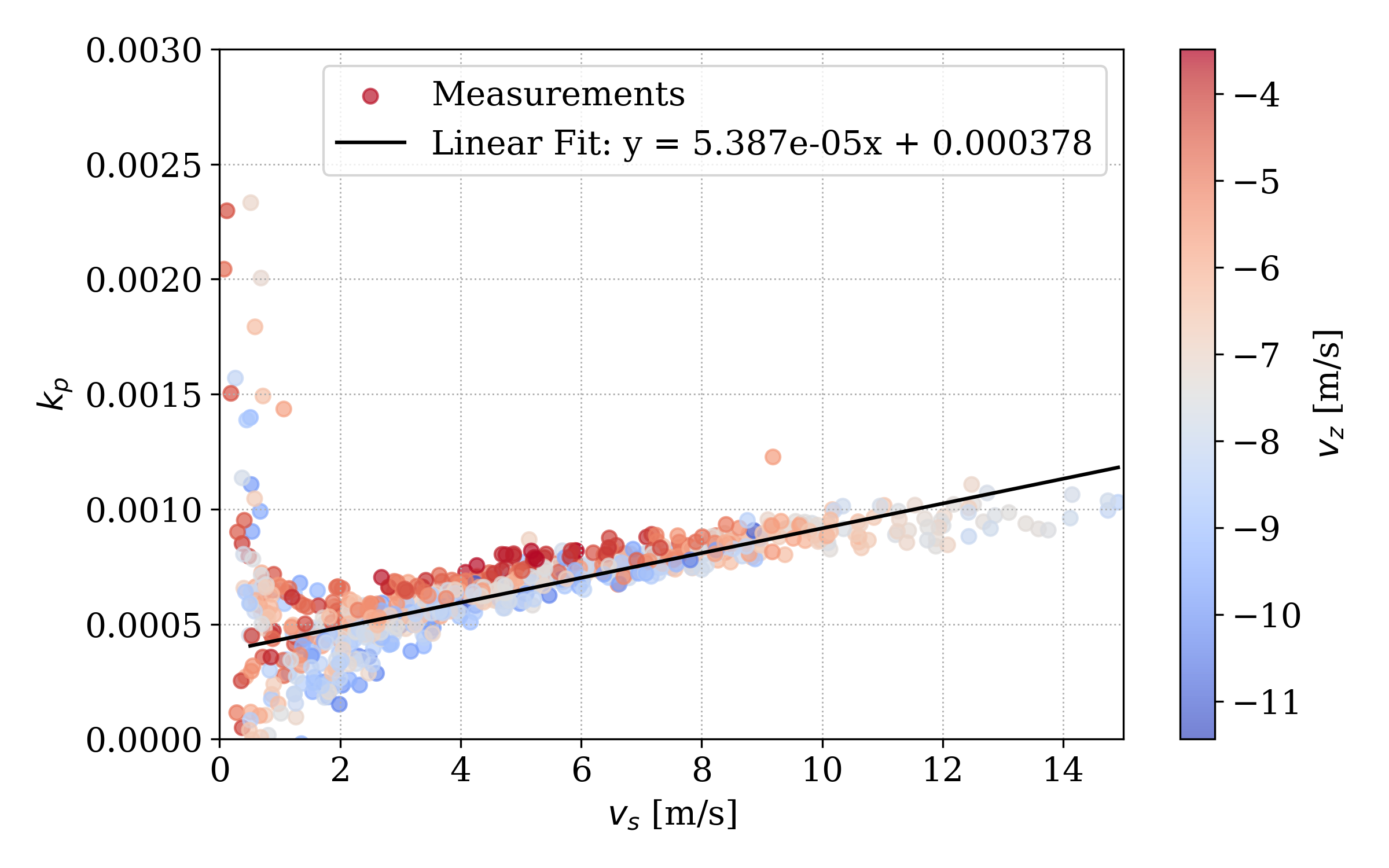}
}

\subfloat[Medium pips (7)]{%
    \includegraphics[width=0.48\linewidth]{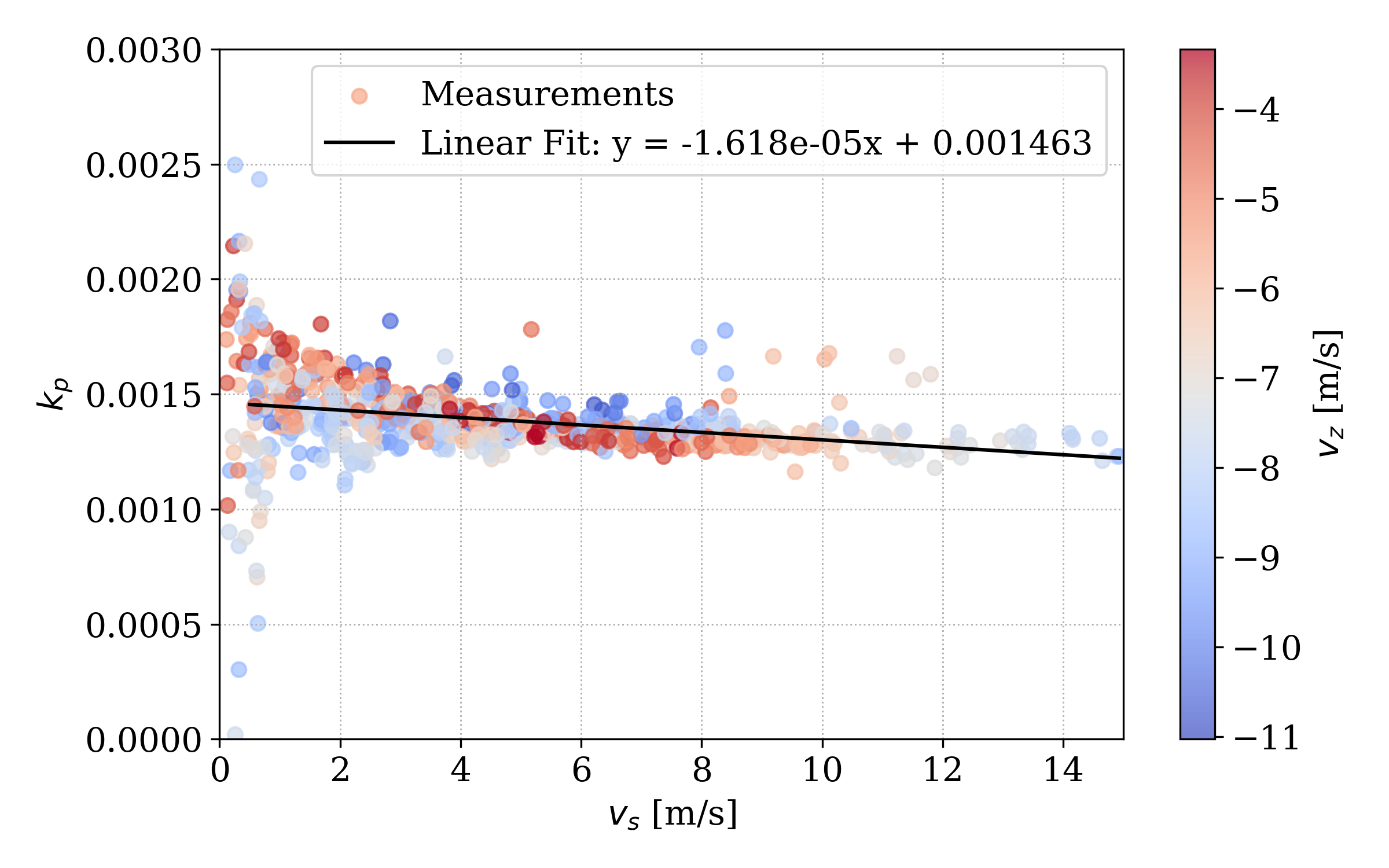}
}
\hfill
\subfloat[Short pips (8)]{%
    \includegraphics[width=0.48\linewidth]{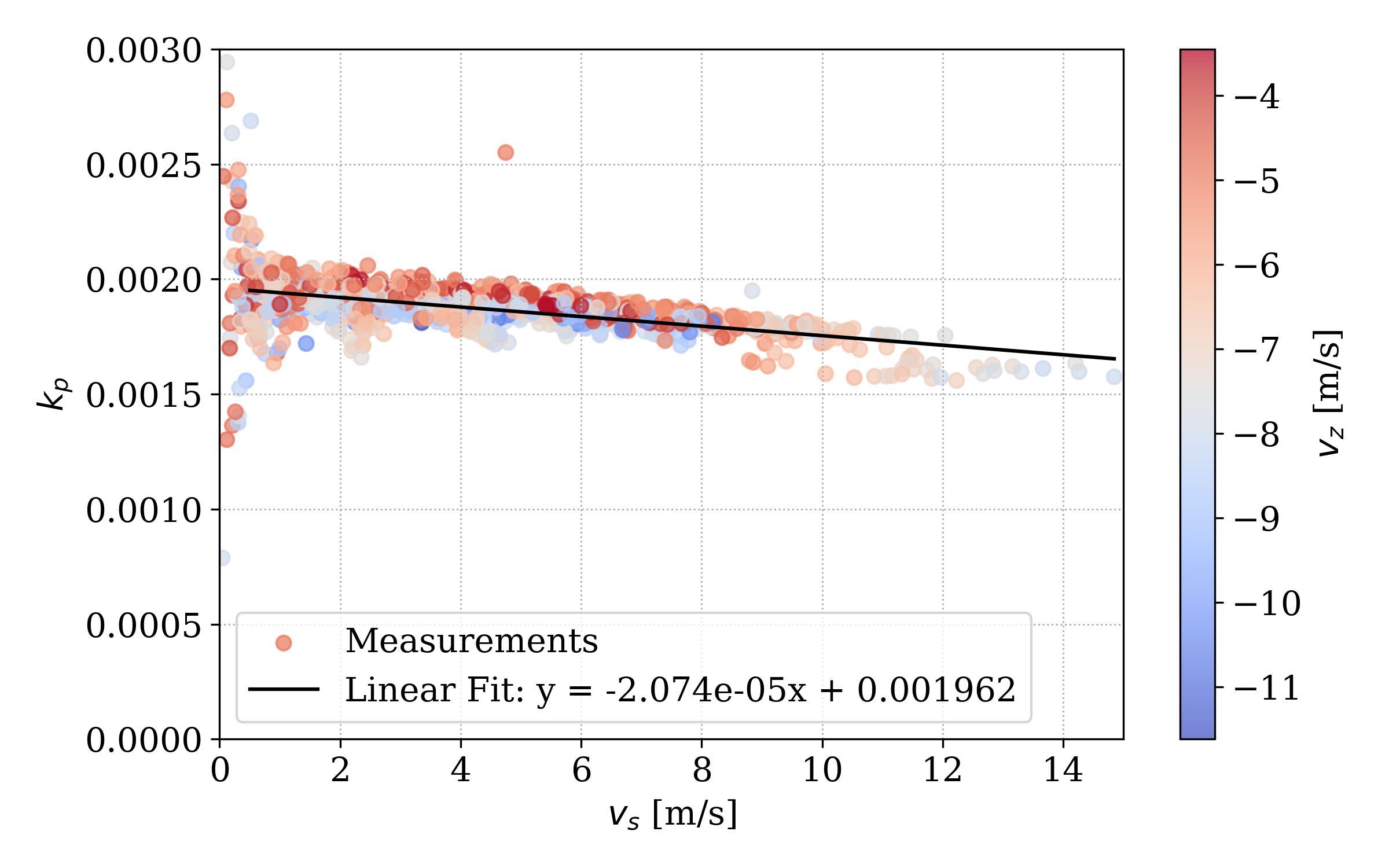}
}

\caption{Tangential stiffness parameter $k_p$ of rackets equipped with pimpled rubbers.}
\label{fig:kp_pip_rubbers}
\end{figure*}

Similarly to the inverted rubbers, we plot $k_p$ as a function of tangential surface velocity $v_s$ in \Cref{fig:kp_pip_rubbers}.
The behavior of $k_p$ for pip rubbers is significantly more complex.
For the long pips (rackets 5 and 6), $k_p$ is much lower than for the other rackets, reflecting their lower sensitivity to spin and their ability to reverse it.
As with the \ac{COR}, $k_p$ depends on both $v_s$ and $v_z$, which is not the case for the short and medium pips (rackets 7 and 8).
In the long pips with sponge (racket 6), we observe a saturation of $k_p$ beyond $v_s \approx 7,\mathrm{m/s}$, likely due to the pips reaching their maximum deformation.
This saturation is not visible in the long pips without sponge (racket 7), possibly because the velocities required to reach this regime were not achieved during testing.
Consequently, a simple linear model is insufficient to describe $k_p$ for long pips.

\begin{figure}[]
\centering

\subfloat[COR]{%
    \includegraphics[width=0.48\linewidth]{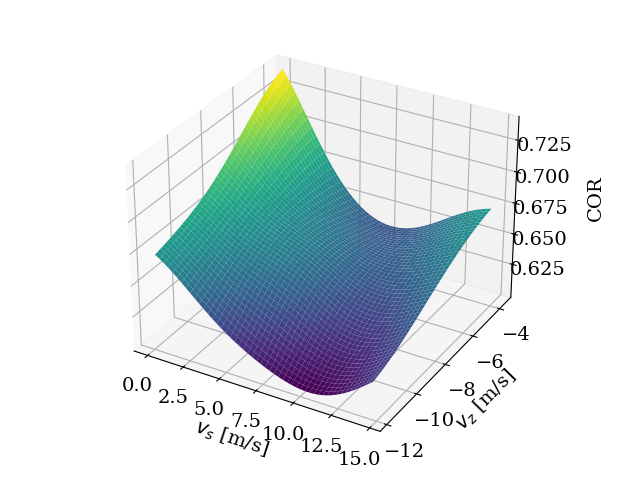}
}
\hfill
\subfloat[$\alpha$]{%
    \includegraphics[width=0.48\linewidth]{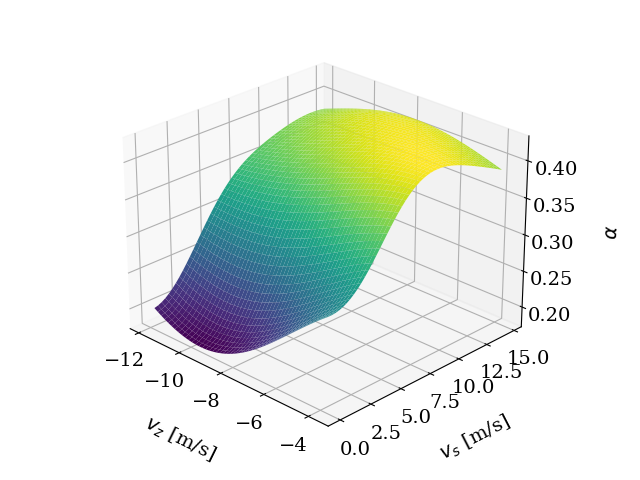}
}

\caption{$\mathrm{COR}$ and $\alpha$ as Gaussian process functions of $v_s$ and $v_z$ for racket~5 (long pips with rubber).}
\label{fig:gp_racket_05}
\end{figure}

For medium and short pips, $|v_z|$ appears to have little effect on the mean value of $k_p$, though it reduces its variance.
%
%
However, this approximation is insufficient to capture the strongly nonlinear behavior of long pimple rubbers.
As shown in \Cref{fig:gp_racket_05}, the mean predictions of the estimated \ac{GP} for $e$ and $\alpha$ (Racket~5) exhibit pronounced nonlinear variations that are not well represented by linear models.

\subsection{General model racket dynamics}

From the analysis of the different racket configurations, several key observations can be drawn.
First, the \ac{COR} is not constant but decreases with increasing impact velocity, reflecting greater energy dissipation at higher speeds.
Similarly, the tangential parameter $k_p$ varies with the tangential velocity, indicating increased energy dissipation for larger surface velocities.
Most notably, long-pimple rubbers exhibit a strong dependence on the tangential velocity, as it directly affects the deformation of the pips and leads to highly nonlinear interaction dynamics.

These observations highlight the limitations of the linear bounce model in \Cref{eq:linear_bounce_model}, which assumes constant coefficients.
In practice, the model parameters vary with the impact conditions, revealing the inherently nonlinear nature of racket-ball interactions.
To better capture these effects while preserving physical interpretability, we reformulate the bounce dynamics as:
\begin{equation}
\begin{split}
\bm{v^{+}} & = \bm{A}(\bm{v_s}, \bm{v_z}) \bm{v^{-}} + \bm{B}(\bm{v_s}, \bm{v_z}) \bm{\omega^{-}}\\
\bm{\omega^{+}} & = \bm{C}(\bm{v_s}, \bm{v_z}) \bm{v^{-}} + \bm{D}(\bm{v_s}, \bm{v_z}) \bm{\omega^{-}}
\end{split}
\label{eq:state_dep_linear_bounce_model}
\end{equation}
For anti-spin and inverted rubbers, the parameter variations can be approximated using simple parametric functions of the tangential and normal velocities ($v_s$, $v_z$).
However, such models are insufficient for pimpled rubbers.
As such, estimating these state-dependent parameters requires a highly flexible model.
While neural network-based approaches~\cite{huang2011b,zhao2016} provide such expressiveness, they typically lack uncertainty estimates and are less amenable to online updates.
To overcome these limitations, we propose modeling parameter variations using \acp{GP}.
\acp{GP} can capture complex nonlinear dependencies while maintaining a principled probabilistic formulation.
In particular, they provide uncertainty estimates that are critical for downstream tasks such as uncertainty-aware trajectory prediction and domain adaptation in reinforcement learning.
Furthermore, their Bayesian nature enables consistent online updates as new data becomes available, without discarding prior information.

\subsection{Benchmark}
\label{ssec:benchmark}
We evaluate the predictive performance of different methods for estimating the bounce parameters and report the results in \Cref{tab:racket_model_eval}.
We measure the prediction error on the ball's post-impact velocity and spin.
Since the observations are restricted to a plane, the velocity error is defined as
\begin{equation}
v_{\mathrm{MAE}} = \frac{1}{N} \sum_{i=1}^{N}
\left\|
\begin{bmatrix}
v^{+}_{x,i} \\
v^{+}_{z,i}
\end{bmatrix}
-
\begin{bmatrix}
\hat{v}^{+}_{x,i} \\
\hat{v}^{+}_{z,i}
\end{bmatrix}
\right\|_2
\end{equation}
and the spin error as
\begin{equation}
\omega_{\mathrm{MAE}} = \frac{1}{N} \sum_{i=1}^{N} \left\| \boldsymbol{\omega}^+_{y,i} - \hat{\boldsymbol{\omega}}^+_{y,i} \right\|_2,
\end{equation}
where $\hat{(\cdot)}$ are the predicted values.

We compare prediction error for models with constant parameters, linear state-dependent parameters, and parameters estimated from $(v_s, v_z)$ using an \ac{MLP}, a residual \ac{MLP}, and a \ac{GP}.
The \ac{MLP}, following \cite{zhao2016}, consists of a single hidden layer with 32 ReLU units that estimates the physical parameters.
The residual \ac{MLP} builds upon the constant-parameter model by learning a correction to be applied to the prediction $(v^+_x, v_z^+, w_y^+)$.
For both models, inputs and outputs are normalized, and weights are regularized with a factor of $10^{-4}$.

As it can be seen in \Cref{tab:racket_model_eval}, the model with constant physical parameters yields the highest errors across all rackets, confirming that a static parameterization is insufficient to capture the variability of real-world bounce dynamics.

Introducing state-dependent parameters as functions of the incoming vertical and tangential velocities $(v_z, v_s)$ leads to a clear improvement in prediction accuracy. 
On average, this reduces the velocity MAE by approximately 12\,cm/s across all racket configurations, highlighting the importance of accounting for impact-dependent effects.
However, this parametric approach remains limited for rackets with long pips (Rackets 5 and 6), where the error reduction is smaller. 
This is consistent with the highly nonlinear and deformation-driven interaction dynamics of long-pimple rubbers, which cannot be well approximated by simple parametric functions.

In contrast, the \ac{GP} consistently achieves the best overall performance, with an average velocity \ac{MAE} of 19\,cm/s and low spin prediction errors across all rackets. 
Its advantage is particularly pronounced for non-inverted rubbers, where the dynamics exhibit strong nonlinearities. 
This suggests that the non-parametric flexibility of GPs, combined with their smoothness priors, enables them to capture complex, state-dependent variations better while remaining robust in a data-limited regime.

The \ac{MLP} baseline achieves competitive performance, with an average velocity error of 20\,cm/s, but does not consistently outperform simpler models. 
In contrast, the residual \ac{MLP} performs worse than the linear model. 
This can be attributed to the lack of explicit physical structure: while the MLP directly predicts physically meaningful parameters, the residual model must implicitly learn both the underlying structure and the correction from limited data. 
As a result, it struggles to generalize and fails to exploit the known physical relationships governing the bounce dynamics fully.

Overall, these results show that state-dependent parameterizations are essential for accurate bounce modeling. 
While simple parametric models improve over constant ones, they remain limited for strongly nonlinear regimes. 
\ac{GP} provide the best performance, as their prior smoothness implicitly regularizes the problem, enabling accurate and robust predictions even in data-limited settings.

\definecolor{bestcolor}{RGB}{198,239,206}    
\definecolor{secondcolor}{RGB}{255,235,156}  
\definecolor{thirdcolor}{RGB}{244,204,204}   

\sisetup{
	separate-uncertainty = true,
	table-align-uncertainty = true,
	detect-all
}

\begin{table*}[ht]
	\centering
	\begin{tabular}{
		c
		>{$}c<{$} >{$}c<{$}
		>{$}c<{$} >{$}c<{$}
		>{$}c<{$} >{$}c<{$}
		>{$}c<{$} >{$}c<{$}
		>{$}c<{$} >{$}c<{$}
	}
	\toprule
	\multirow{2}{*}{Racket} &
	\multicolumn{2}{c}{\textbf{Constant}} &
	\multicolumn{2}{c}{\textbf{Linear}} &
	\multicolumn{2}{c}{\textbf{MLP}} &
	\multicolumn{2}{c}{\textbf{Residual MLP}} &
	\multicolumn{2}{c}{\textbf{GP Model}} \\
	& \text{MAE }\bm{v'} & \text{MAE }\bm{\omega'}
	& \text{MAE }\bm{v'} & \text{MAE }\bm{\omega'}
	& \text{MAE }\bm{v'} & \text{MAE }\bm{\omega'}
	& \text{MAE }\bm{v'} & \text{MAE }\bm{\omega'}
	& \text{MAE }\bm{v'} & \text{MAE }\bm{\omega'} \\
	\midrule
	1  & 33 \pm 23
	   & 14 \pm 15
	   & \cellcolor{secondcolor}21 \pm 13
	   & 12 \pm 8
	   & \cellcolor{secondcolor}21 \pm 13
	   & \cellcolor{thirdcolor}11 \pm 8
	   & \cellcolor{secondcolor}21 \pm 15
	   & \cellcolor{bestcolor}8 \pm 9
	   & \cellcolor{bestcolor}20 \pm 12
	   & \cellcolor{thirdcolor}11 \pm 7 \\
	2  & 27 \pm 22
	   & 8 \pm 9
	   & \cellcolor{thirdcolor}20 \pm 13
	   & 8 \pm 7
	   & \cellcolor{secondcolor}19 \pm 13
	   & \cellcolor{thirdcolor}7 \pm 6
	   & 20 \pm 14
	   & \cellcolor{thirdcolor}7 \pm 6
	   & \cellcolor{bestcolor}18 \pm 13
	   & \cellcolor{bestcolor}6 \pm 5 \\
	3  & 37 \pm 28
	   & 16 \pm 18
	   & \cellcolor{secondcolor}24 \pm 17
	   & 13 \pm 9
	   & \cellcolor{secondcolor}24 \pm 17
	   & 13 \pm 10
	   & \cellcolor{thirdcolor}25 \pm 19
	   & \cellcolor{bestcolor}11 \pm 13
	   & \cellcolor{bestcolor}23 \pm 17
	   & \cellcolor{bestcolor}11 \pm 9 \\
	4  & 30 \pm 22
	   & 9 \pm 11
	   & 20 \pm 13
	   & 8 \pm 7
	   & \cellcolor{bestcolor}18 \pm 13
	   & \cellcolor{thirdcolor}7 \pm 6
	   & 20 \pm 15
	   & \cellcolor{bestcolor}6 \pm 7
	   & \cellcolor{bestcolor}18 \pm 13
	   & \cellcolor{thirdcolor}7 \pm 6 \\
	5  & 48 \pm 45
	   & 27 \pm 28
	   & \cellcolor{thirdcolor}27 \pm 22
	   & 13 \pm 15
	   & \cellcolor{secondcolor}24 \pm 19
	   & \cellcolor{thirdcolor}11 \pm 12
	   & 41 \pm 37
	   & 24 \pm 24
	   & \cellcolor{bestcolor}19 \pm 17
	   & \cellcolor{bestcolor}7 \pm 8 \\
	6  & 43 \pm 38
	   & 25 \pm 30
	   & \cellcolor{thirdcolor}25 \pm 14
	   & 11 \pm 13
	   & \cellcolor{bestcolor}20 \pm 13
	   & \cellcolor{bestcolor}8 \pm 12
	   & 35 \pm 33
	   & 21 \pm 27
	   & \cellcolor{bestcolor}20 \pm 13
	   & \cellcolor{secondcolor}9 \pm 12 \\
	7  & 26 \pm 20
	   & 12 \pm 13
	   & \cellcolor{thirdcolor}19 \pm 18
	   & \cellcolor{bestcolor}8 \pm 9
	   & \cellcolor{bestcolor}18 \pm 18
	   & \cellcolor{bestcolor}8 \pm 9
	   & 20 \pm 17
	   & 11 \pm 12
	   & \cellcolor{bestcolor}18 \pm 17
	   & \cellcolor{bestcolor}8 \pm 10 \\
	8  & 27 \pm 25
	   & 12 \pm 17
	   & \cellcolor{bestcolor}17 \pm 19
	   & 11 \pm 9
	   & \cellcolor{bestcolor}17 \pm 19
	   & \cellcolor{secondcolor}10 \pm 8
	   & 21 \pm 22
	   & \cellcolor{secondcolor}10 \pm 12
	   & \cellcolor{bestcolor}17 \pm 18
	   & \cellcolor{bestcolor}9 \pm 7 \\
	9  & 30 \pm 23
	   & 15 \pm 19
	   & \cellcolor{thirdcolor}17 \pm 13
	   & 13 \pm 9
	   & \cellcolor{bestcolor}16 \pm 13
	   & \cellcolor{thirdcolor}11 \pm 7
	   & 18 \pm 13
	   & \cellcolor{bestcolor}8 \pm 8
	   & \cellcolor{bestcolor}16 \pm 13
	   & \cellcolor{thirdcolor}11 \pm 7 \\
	10 & 27 \pm 18
	   & 16 \pm 14
	   & \cellcolor{secondcolor}23 \pm 15
	   & \cellcolor{secondcolor}9 \pm 9
	   & 26 \pm 14
	   & 12 \pm 9
	   & \cellcolor{thirdcolor}25 \pm 16
	   & 13 \pm 11
	   & \cellcolor{bestcolor}22 \pm 13
	   & \cellcolor{bestcolor}8 \pm 7 \\
	\midrule
	\rowcolor{gray!15}
	\textbf{Mean} &
	33 \pm 26 &
	15 \pm 17 &
	\cellcolor{thirdcolor}21 \pm 16 &
	\cellcolor{thirdcolor}11 \pm 9 &
	\cellcolor{secondcolor} 20 \pm 15 &
	\cellcolor{secondcolor} 10 \pm 9 &
	25 \pm 20 &
	12 \pm 13 &
	\cellcolor{bestcolor}\mathbf{19 \pm 15} &
	\cellcolor{bestcolor}\mathbf{9 \pm 8} \\
	\bottomrule
	\end{tabular}
    \vspace{2mm}
	\caption{Evaluation of table tennis racket models with different bounce models. All methods use the same physical model and differ only in how the bounce parameters are estimated. Reported results correspond to the 2D variants of the learned models. For each racket and metric, the best, second-best, and third-best results are highlighted in green, yellow, and red, respectively.}
	\label{tab:racket_model_eval}
\end{table*}

\section{Online Adaptation}
\label{sec:online_adaptation}
During a table tennis rally, the opponent's racket properties are initially unknown, and only a coarse prior can be assumed.
Through repeated interactions, players progressively refine their understanding of the bounce dynamics and improve their predictions.
We aim to equip the robot with a similar capability.

Naively performing online system identification would require a large number of samples and would fail to leverage prior knowledge about racket dynamics.
Similarly, fine-tuning an \ac{MLP} online is computationally demanding, prone to overfitting on limited observations, and may generalize poorly to unseen ball states.

We therefore formulate online adaptation as Bayesian inference using \acp{GP}: starting from a prior over racket dynamics, the model is updated with observed ball--racket interactions during the rally to obtain a posterior model that is both data-efficient and uncertainty-aware.
The prior is constructed from training data of rackets sharing the same rubber type.
While exact GP training scales as $O(N^3)$ in time and $O(N^2)$ in memory, this step is performed offline.
At test time, inference with an RBF kernel scales as $O(dN)$ for the mean and up to $O(N^2)$ for the variance, which can become too expensive for real-time use when many queries are required.

To enable online use, we replace the exact \ac{RBF} prior with a linear prior defined on the features $(1, v_s, v_z)$.
This choice provides a favorable trade-off between efficiency and accuracy, as the bounce parameters of inverted rubbers are already well approximated by linear models, as shown in \Cref{ssec:benchmark}.
In this case, inference reduces to a low-dimensional parametric model, independent of the dataset size.
We evaluate this approach on inverted rubbers, using rackets 1, 2, 3, 4, and 9 to estimate the prior.
To account for inter-racket variability, we define a conservative prior variance as $\sigma_{\text{prior}}=\sigma_f+\sigma_n$.

\begin{figure}[]
    \centering
    \includegraphics[width=\linewidth]{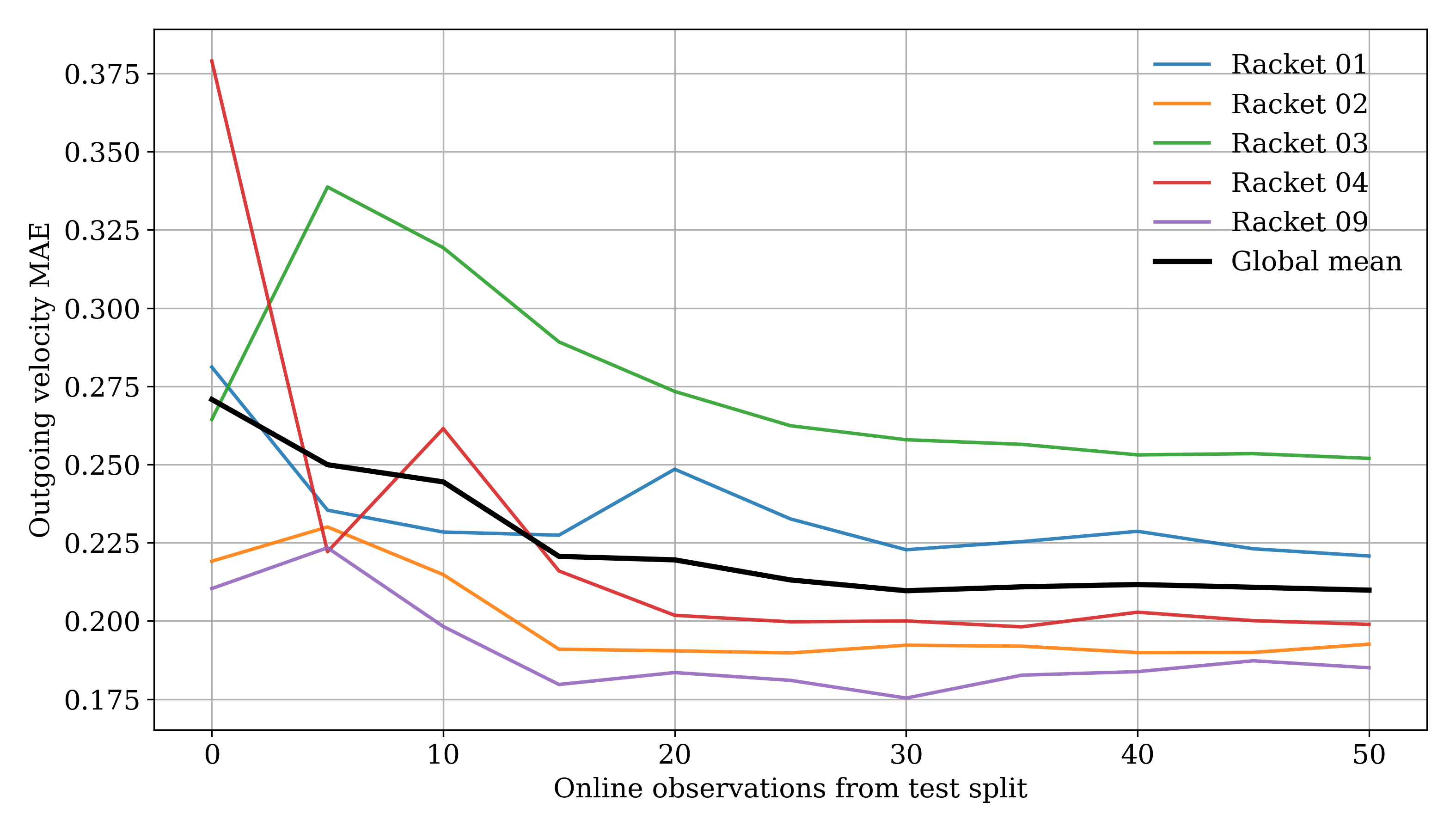}
    \caption{
    Evolution of the velocity prediction error (MAE) during online adaptation with regards to the number of observations. Prior is defined with training samples, while observed samples are taken from the test set.
    }
    \label{fig:posterior_vel_error}
\end{figure}

\Cref{fig:posterior_vel_error} shows the evolution of the velocity prediction error during adaptation.
After a short burn-in phase of about 10 samples, the posterior model consistently improves and reduces the velocity prediction error after roughly 30 observations.
Overall, the mean error decreases by about 6~cm/s, and by up to 17~cm/s in the most out-of-distribution cases.

\begin{figure}[]
    \centering
    \includegraphics[width=\linewidth]{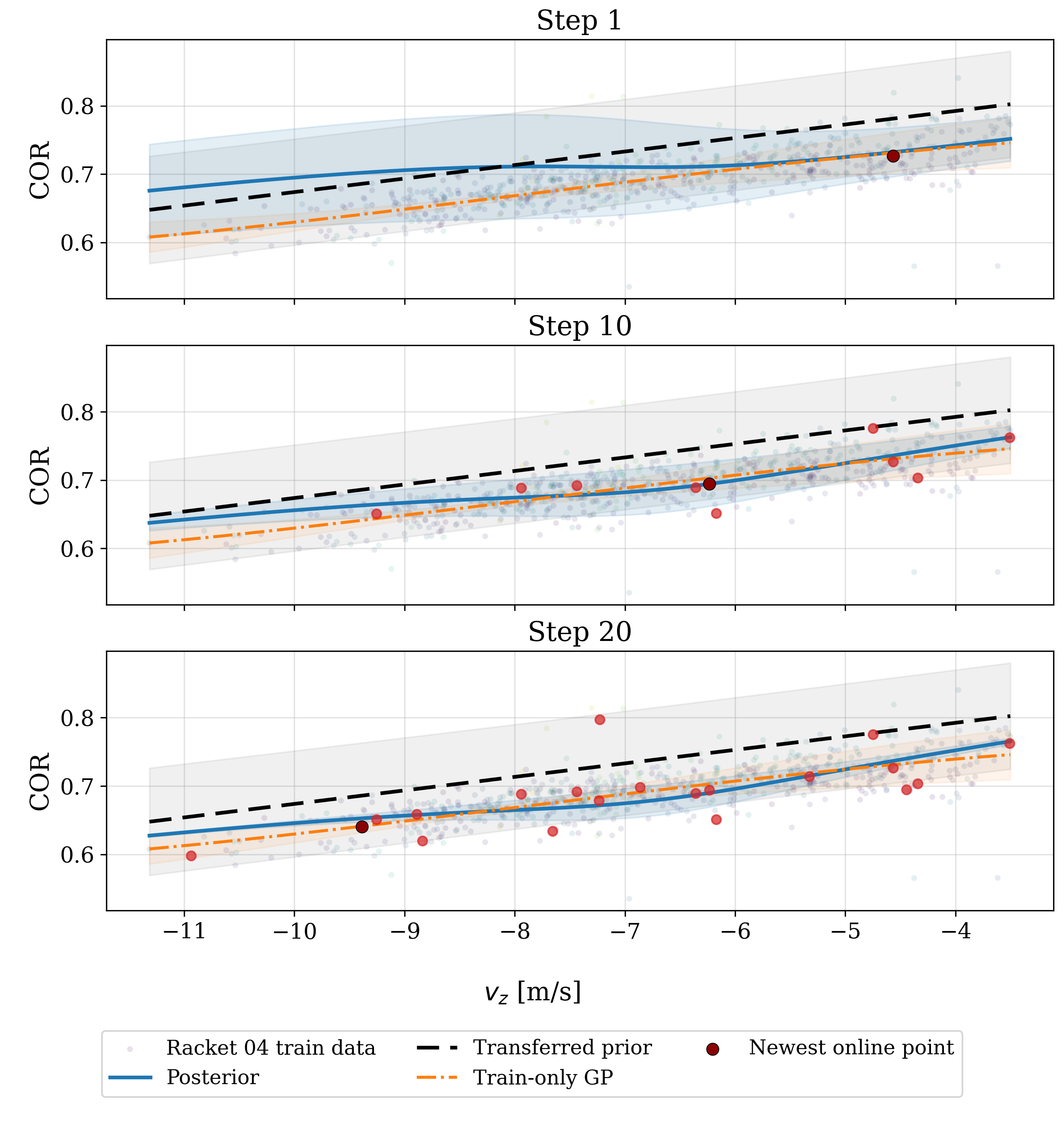}
    \caption{
    Evolution of the posterior estimate of the coefficient of restitution (COR) for racket~04 during online adaptation. 
    The model is updated with 1, 10, and 20 observations. 
    }
    \label{fig:racket_posterior_evolution}
\end{figure}

\Cref{fig:racket_posterior_evolution} illustrates this adaptation process for a representative racket.
As more samples are incorporated, the posterior progressively departs from the transferred prior and better matches the racket-specific dynamics, while reducing uncertainty.

\section{Limitations}
\label{sec:limitations}
Despite the strong predictive performance of the proposed model, several limitations remain.
First, the experimental setup constrains the ball motion to a plane, implicitly assuming isotropic bounce behavior in the tangential directions, which may not fully capture the true three-dimensional dynamics of real impacts.
Second, the dataset spans a limited range of velocities and spin magnitudes, constrained by the capabilities of the ball launcher, which may limit the model's ability to generalize to more extreme or professional-level playing conditions.
Finally, the proposed approach relies on an impulse-based contact model that assumes instantaneous interactions and may not capture prolonged contact effects.
This is particularly limiting for strokes such as serves, where increased contact time is used to maximize spin transfer.

\section{Conclusion}
\label{sec:conclusion}
We presented a general framework for modeling ball-racket interactions in robotic table tennis across a wide range of rubber types.
Our analysis showed that commonly used constant-parameter contact models fail to capture the strongly state-dependent rebound behavior observed in real impacts, particularly for pimpled rubbers.
To address this limitation, we modeled the parameters of an impulse-based bounce model as functions of the incoming ball state using \ac{GP}, preserving physical structure while capturing nonlinear variations and providing uncertainty estimates.
The resulting model achieves improved prediction accuracy of post-impact velocity and spin compared to constant and parametric baselines, with the largest gains observed for nonstandard rubbers, and enables reliable target-directed hit planning.
Future work will focus on integrating the proposed model into control frameworks for improved robotic table tennis performance.

\bibliographystyle{IEEEtran}
\bibliography{biblio}

@inproceedings{bao2012,
  title = {Bouncing Model for the Table Tennis Trajectory Prediction and the Strategy of Hitting the Ball},
  booktitle = {2012 {{IEEE International Conference}} on {{Mechatronics}} and {{Automation}}},
  author = {Bao, Han and Chen, Xiaopeng and Wang, Zhan Tao and Pan, Min and Meng, Fei},
  year = 2012,
  month = aug,
  pages = {2002--2006},
  issn = {2152-744X},
  doi = {10.1109/ICMA.2012.6285129},
  urldate = {2025-01-02}
}

@book{bishop2006,
  title = {Pattern {{Recognition}} and {{Machine Learning}}},
  author = {Bishop, Christopher},
  year = 2006,
  volume = {4},
  publisher = {Springer, New York}
}

@article{care2023,
  title = {Kernel {{Methods}} and {{Gaussian Processes}} for {{System Identification}} and {{Control}}: {{A Road Map}} on {{Regularized Kernel-Based Learning}} for {{Control}}},
  shorttitle = {Kernel {{Methods}} and {{Gaussian Processes}} for {{System Identification}} and {{Control}}},
  author = {Car{\`e}, Algo and Carli, Ruggero and Libera, Alberto Dalla and Romeres, Diego and Pillonetto, Gianluigi},
  year = 2023,
  month = oct,
  journal = {IEEE Control Systems Magazine},
  volume = {43},
  number = {5},
  pages = {69--110},
  issn = {1941-000X},
  doi = {10.1109/MCS.2023.3291625},
  urldate = {2026-01-07}
}

@article{chen2010,
  title = {Dynamic Model Based Ball Trajectory Prediction for a Robot Ping-Pong Player},
  author = {Chen, Xiaopeng and Tian, Ye and Huang, Qiang and Zhang, Weimin and Yu, Zhangguo},
  year = 2010,
  month = dec,
  journal = {2010 IEEE International Conference on Robotics and Biomimetics, ROBIO 2010},
  doi = {10.1109/ROBIO.2010.5723394}
}

@inproceedings{dambrosio2024,
  title = {Achieving Human Level Competitive Robot Table Tennis},
  booktitle = {7th Robot Learning Workshop: {{Towards}} Robots with Human-Level Abilities},
  author = {D'Ambrosio, David B and Abeyruwan, Saminda Wishwajith and Graesser, Laura and Iscen, Atil and Amor, Heni Ben and Bewley, Alex and Reed, Barney and Reymann, Krista and Takayama, Leila and Tassa, Yuval and others},
  year = 2024
}

@inproceedings{Gossard2023iros,
  title = {{{SpinDOE}}: A Ball Spin Estimation Method for Table Tennis Robot},
  booktitle = {2023 {{IEEE}}/{{RSJ}} International Conference on Intelligent Robots and Systems ({{IROS}})},
  author = {Gossard, Thomas and Tebbe, Jonas and Ziegler, Andreas and Zell, Andreas},
  year = 2023,
  month = oct,
  publisher = {IEEE},
  doi = {10.1109/iros55552.2023.10342178}
}

@article{hayakawa2016,
  title = {Ball {{Trajectory Planning}} in {{Serving Task}} for {{Table Tennis Robot}}},
  author = {Hayakawa, Yoshikazu and Nakashima, Akira and Itoh, Satoshi and Nakai, Yuki},
  year = 2016,
  month = mar,
  journal = {SICE Journal of Control, Measurement, and System Integration},
  volume = {9},
  number = {2},
  pages = {50--59},
  issn = {1882-4889, 1884-9970},
  doi = {10.9746/jcmsi.9.50},
  urldate = {2026-01-07},
  langid = {english}
}

@inproceedings{huang2011a,
  title = {Trajectory Prediction of Spinning Ball for Ping-Pong Player Robot},
  booktitle = {{{IEEE International Conference}} on {{Intelligent Robots}} and {{Systems}}},
  author = {Huang, Yanlong and Xu, De and Tan, M. and Su, Hu},
  year = 2011,
  month = sep,
  pages = {3434--3439},
  doi = {10.1109/IROS.2011.6095044}
}

@article{huang2011b,
  title = {Improved {{Detection}} of {{Ball Hit Events}} in a {{Tennis Game Using Multimodal Information}}},
  author = {Huang, Qiang and Cox, Stephen and Yan, Fei and {de Campos}, Teo and Windridge, David and Kittler, Josef and Christmas, William},
  year = 2011,
  langid = {english}
}

@inproceedings{kyohei2019,
  title = {The {{Ping Pong Robot}} to {{Return}} a {{Ball Precisely}} \textasciitilde{} {{Trajectory Prediction}} and {{Racket Control}} for {{Spinning Balls}}},
  author = {Kyohei, Asai and Masamune, Nakayama and Satoshi, Yase},
  year = 2019,
  urldate = {2026-01-01}
}

@inproceedings{liu2012,
  title = {Racket Control and Its Experiments for Robot Playing Table Tennis},
  booktitle = {2012 {{IEEE International Conference}} on {{Robotics}} and {{Biomimetics}} ({{ROBIO}})},
  author = {Liu, Chunfang and Hayakawa, Yoshikazu and Nakashima, Akira},
  year = 2012,
  month = dec,
  pages = {241--246},
  doi = {10.1109/ROBIO.2012.6490973},
  urldate = {2026-01-25}
}

@article{liu2024,
  title = {Physics-Informed Neural Networks to Model and Control Robots: {{A}} Theoretical and Experimental Investigation},
  author = {Liu, Jingyue and Borja, Pablo and Della Santina, Cosimo},
  year = 2024,
  journal = {Advanced Intelligent Systems},
  volume = {6},
  number = {5},
  pages = {2300385}
}

@article{muratore2021a,
  title = {Data-{{Efficient Domain Randomization With Bayesian Optimization}}},
  author = {Muratore, Fabio and Eilers, Christian and Gienger, Michael and Peters, Jan},
  year = 2021,
  month = apr,
  journal = {IEEE Robotics and Automation Letters},
  volume = {6},
  number = {2},
  pages = {911--918},
  issn = {2377-3766},
  doi = {10.1109/LRA.2021.3052391},
  urldate = {2026-01-11}
}

@inproceedings{nakashima2010,
  title = {Modeling of Rebound Phenomenon of a Rigid Ball with Friction and Elastic Effects},
  booktitle = {Proceedings of the 2010 {{American Control Conference}}},
  author = {Nakashima, A and Ogawa, Y and Kobayashi, Y and Hayakawa, Y},
  year = 2010,
  month = jun,
  pages = {1410--1415},
  publisher = {IEEE},
  address = {Baltimore, MD},
  doi = {10.1109/ACC.2010.5530520},
  urldate = {2023-11-14},
  isbn = {978-1-4244-7427-1 978-1-4244-7426-4 978-1-4244-7425-7},
  langid = {english}
}

@inproceedings{nakashima2011,
  title = {Robotic Table Tennis Based on Physical Models of Aerodynamics and Rebounds},
  booktitle = {2011 {{IEEE International Conference}} on {{Robotics}} and {{Biomimetics}}},
  author = {Nakashima, Akira and Ogawa, Yuki and Liu, Chunfang and Hayakawa, Yoshikazu},
  year = 2011,
  month = dec,
  pages = {2348--2354},
  doi = {10.1109/ROBIO.2011.6181649},
  urldate = {2026-01-25}
}

@misc{nguyen2025b,
  title = {Whole {{Body Model Predictive Control}} for {{Spin-Aware Quadrupedal Table Tennis}}},
  author = {Nguyen, David and Zaidi, Zulfiqar and Karol, Kevin and Hodgins, Jessica and Xie, Zhaoming},
  year = 2025,
  month = oct,
  number = {arXiv:2510.08754},
  eprint = {2510.08754},
  primaryclass = {cs},
  publisher = {arXiv},
  doi = {10.48550/arXiv.2510.08754},
  urldate = {2026-01-25},
  archiveprefix = {arXiv}
}

@incollection{sarkka2019,
  title = {The {{Use}} of {{Gaussian Processes}} in {{System Identification}}},
  booktitle = {Encyclopedia of {{Systems}} and {{Control}}},
  author = {S{\"a}rkk{\"a}, Simo},
  year = 2019,
  pages = {1--10},
  publisher = {Springer, London},
  doi = {10.1007/978-1-4471-5102-9_100087-1},
  urldate = {2026-01-07},
  isbn = {978-1-4471-5102-9},
  langid = {english}
}

@article{yang2023a,
  title = {The Effect of Aerodynamics on Table Tennis},
  author = {Yang, Qinghua},
  year = 2023,
  month = may,
  journal = {Theoretical and Natural Science},
  volume = {5},
  pages = {465--473},
  doi = {10.54254/2753-8818/5/20230284}
}

@article{zhao2016,
  title = {Rebound {{Modeling}} of {{Spinning Ping-Pong Ball Based}} on {{Multiple Visual Measurements}}},
  author = {Zhao, Yongsheng and Xiong, Rong and Zhang, Yifeng},
  year = 2016,
  month = aug,
  journal = {IEEE Transactions on Instrumentation and Measurement},
  volume = {65},
  pages = {1--11},
  doi = {10.1109/TIM.2016.2555179}
}

@article{zhao2017a,
  title = {Model {{Based Motion State Estimation}} and {{Trajectory Prediction}} of {{Spinning Ball}} for {{Ping-Pong Robots}} Using {{Expectation-Maximization Algorithm}}},
  author = {Zhao, Yongsheng and Xiong, Rong and Zhang, Yifeng},
  year = 2017,
  month = sep,
  journal = {Journal of Intelligent \& Robotic Systems},
  volume = {87},
  number = {3},
  pages = {407--423},
  issn = {1573-0409},
  doi = {10.1007/s10846-017-0515-8},
  urldate = {2026-01-07},
  langid = {english}
}

@book{kocijan2004gaussian,
  title={Gaussian process model based predictive control},
  author={Kocijan, Jus and Murray-Smith, Roderick and Rasmussen, Carl E and Girard, Agathe},
  volume={3},
  year={2004},
  publisher={Institute of Electrical and Electronics Engineers (IEEE)}
}

@article{hewing2019cautious,
  title={Cautious model predictive control using gaussian process regression},
  author={Hewing, Lukas and Kabzan, Juraj and Zeilinger, Melanie N},
  journal={IEEE Transactions on Control Systems Technology},
  volume={28},
  number={6},
  pages={2736--2743},
  year={2019},
  publisher={IEEE}
}

@article{quarteroni2025combining,
  title={Combining physics-based and data-driven models: advancing the frontiers of research with scientific machine learning},
  author={Quarteroni, Alfio and Gervasio, Paola and Regazzoni, Francesco},
  journal={arXiv preprint arXiv:2501.18708},
  year={2025}
}

@inproceedings{turner2009system,
  title={System identification in Gaussian process dynamical systems},
  author={Turner, Ryan and Deisenroth, Marc Peter and Rasmussen, Carl Edward},
  booktitle={Nonparametric Bayes Workshop (NIPS 2009), Whistler, BC, Canada},
  year={2009}
}

@inproceedings{buisson2020actively,
  title={Actively learning gaussian process dynamics},
  author={Buisson-Fenet, Mona and Solowjow, Friedrich and Trimpe, Sebastian},
  booktitle={Learning for dynamics and control},
  pages={5--15},
  year={2020},
  organization={PMLR}
}

@inproceedings{nagabandi2018neural,
  title={Neural network dynamics for model-based deep reinforcement learning with model-free fine-tuning},
  author={Nagabandi, Anusha and Kahn, Gregory and Fearing, Ronald S and Levine, Sergey},
  booktitle={2018 IEEE international conference on robotics and automation (ICRA)},
  pages={7559--7566},
  year={2018},
  organization={IEEE}
}

@article{pillonetto2025deep,
  title={Deep networks for system identification: a survey},
  author={Pillonetto, Gianluigi and Aravkin, Aleksandr and Gedon, Daniel and Ljung, Lennart and Ribeiro, Ant{\^o}nio H and Sch{\"o}n, Thomas B},
  journal={Automatica},
  volume={171},
  pages={111907},
  year={2025},
  publisher={Elsevier}
}
\vfill

\end{document}